%% file: Landsat30-AU.tex
\documentclass[letterpaper]{article} 
\usepackage{aaai2026}  
\usepackage{times}  
\usepackage{helvet}  
\usepackage{courier}  
\usepackage[hyphens]{url}  
\usepackage{graphicx} 
\usepackage{amssymb}
\usepackage{xcolor}  
\usepackage{natbib}  
\usepackage{caption} 
\usepackage{algorithm}
\usepackage{algorithmic}
\usepackage{xspace}

\usepackage{booktabs} 
\usepackage{tabularx} 
\usepackage{subcaption} 
\usepackage{multirow} 

\usepackage{newfloat}
\usepackage{listings}
\DeclareCaptionStyle{ruled}{labelfont=normalfont,labelsep=colon,strut=off} 
\floatstyle{ruled}
\newfloat{listing}{tb}{lst}{}
\floatname{listing}{Listing}
\newcommand{\dataset}[1]{\textsc{#1}\xspace}

\nocopyright

\newcommand{\LandsatAU}{\dataset{Landsat30-AU}}

\title{Landsat30-AU: A Vision-Language Dataset for Australian Landsat Imagery}
\author{
  Sai Ma\textsuperscript{\rm 1},
  Zhuang Li\textsuperscript{\rm 2 \thanks{corresponding author}},
  John A.\ Taylor\textsuperscript{\rm 1}
}
\affiliations{
  \textsuperscript{\rm 1}School of Computing, Australian National University, Australia\quad
  \textsuperscript{\rm 2}School of Computing Technologies, Royal Melbourne Institute of Technology, Australia\\
  sai.ma@anu.edu.au, zhuang.li@rmit.edu.au, john.taylor@anu.edu.au
}

\begin{document}

\maketitle

\begin{abstract}

Vision language models (VLMs) that enable natural language interaction with satellite imagery can democratize Earth observation by accelerating expert workflows, making data accessible to non-specialists, and enabling planet-scale automation. However, existing datasets focus mainly on short-term, high-resolution imagery from a limited number of satellites, overlooking low-resolution, multi-satellite, long-term archives, such as Landsat, that are essential for affordable and bias-robust global monitoring. We address this gap with Landsat30-AU, a large-scale vision-language dataset built from 30-meter resolution imagery collected by four Landsat satellites (5, 7, 8, and 9) over Australia, spanning more than 36 years. The dataset includes two components: Landsat30-AU-Cap, containing $196,262$ image-caption pairs, and Landsat30-AU-VQA, comprising 17,725 human-verified visual question answering (VQA) samples across eight remote sensing domains. Both datasets are curated through a bootstrapped pipeline that leverages generic VLMs with iterative refinement and human verification to ensure quality. Our evaluation of eight VLMs on our benchmark reveals that off-the-shelf models struggle to understand satellite imagery. The open-source remote-sensing VLM EarthDial achieves only \textbf{0.07 SPIDEr} in captioning and a VQA accuracy of \textbf{0.48}, highlighting the limitations of current approaches. Encouragingly, lightweight fine-tuning of Qwen2.5-VL-7B on \LandsatAU{} improves captioning performance from \textbf{0.11} to \textbf{0.31 SPIDEr} and boosts VQA accuracy from \textbf{0.74} to \textbf{0.87}.

\end{abstract}

\begin{links}
    \link{Code}{https://github.com/papersubmit1/landsat30-au}
\end{links}

\section{Introduction}

For over fifty years, the \emph{Landsat} program has provided a globally consistent, open-access archive of optical satellite imagery at 30-meter ground-sample distance (GSD) \cite{WULDER2022113195}. Since 1972, eight Landsat satellites have been launched, each equipped with distinct sensors and band configurations, resulting in varying appearances of standard red-green-blue composites across missions \cite{usgs_landsat_band_design}. The upcoming \emph{Landsat Next} series will significantly increase daily acquisition volume, from 900 GB (750 scenes) to 8.2 TB (2,220 scenes), through expanded spectral coverage and improved sensor technology \cite{usgs_landsat_next_news,usgs_landsat_next_data_volume_change}. Meanwhile, vision-language models (VLMs) have shown impressive capabilities in managing and interpreting large-scale Earth observation data, especially with high-resolution sources such as Sentinel-2 imagery \cite{Kuckreja_2024_CVPR,10679571,rs16091477,yuan2024chatearthnetglobalscaleimagetextdataset}. Following these trends, VLMs offer a timely opportunity as natural-language interfaces for long-term, cost-effective, planet-scale analysis using the growing Landsat archive.

Progress is nevertheless constrained by the absence of large-scale image-text corpora that match Landsat’s unique regime. Most existing remote-sensing datasets (i) \textit{focus on sub-meter commercial imagery}, which encourages captions centered on fine-grained objects, such as cars, rooftops, or road markings, that are invisible at 30 m resolution, and often come with restrictive licensing costs that hinder global-scale applications \cite{7546397,ge2025rstellerscalingvisuallanguage};  
(ii) \textit{cover only one or two Landsat satellites}, preventing VLMs from learning the radiometric and band-layout differences that span the full eight-mission Landsat program, and thus limiting their robustness to sensor shifts; and  
(iii) \textit{include Landsat imagery with only a short temporal span}, depriving models of exposure to long-term seasonal patterns, land-cover change, and climate-driven dynamics critical for temporal generalization.  
For example, the datasets that \textit{do} incorporate Landsat imagery fall short: \textsc{EarthDial} includes 1.6 million image patches, but only from Landsat 8 \cite{soni2025earthdialturningmultisensoryearth}, while \textsc{SSL4EO-L} provides five million multi-temporal patches across several missions, yet lacks the associated textual supervision necessary for vision-language alignment \cite{stewart2023ssl4eoldatasetsfoundationmodels}. As a result, there is still no dataset that delivers the long-term, multi-satellite, and resolution-aware supervision needed to develop VLMs for scalable and bias-robust Earth monitoring.

Generating high-quality text annotations for remote sensing images also remains a significant challenge. Manual captioning by domain experts ensures high accuracy \cite{7546397,Zhan_2023} but does not scale. Crowdsourcing or automatic alternatives, such as OpenStreetMap (OSM) \cite{openstreetmap_community} tags or web alt-text \cite{10.1007/978-3-031-72904-1_26,zavras2025gaiaglobalmultimodalmultiscale}, offer scalability but suffer from two key issues: (i) spatial mismatch, where many labeled objects (e.g., \texttt{clinic}, \texttt{cemetery}) are too small to be resolved in 30 m Landsat imagery, and (ii) temporal misalignment, where the metadata may describe a scene years before or after the satellite image was acquired, leading to outdated associations.

To address these limitations, we present \textbf{\LandsatAU{}}, the first large-scale vision-language dataset constructed entirely from 30-meter resolution imagery captured by four Landsat missions (5, 7, 8, and 9) across Australia, spanning from 1988 to 2024. It consists of two parts: (i) \textsc{Landsat30-AU-Cap}, containing $196,262$ image-caption pairs, and (ii) \textsc{Landsat30-AU-VQA}, comprising $17,725$ human-verified visual question answering (VQA) examples covering eight common remote sensing tasks. To address the challenges of scale and label quality, we develop a semi-automatic bootstrapped pipeline that extends the methodologies of \textsc{LHRS-Align} \cite{10.1007/978-3-031-72904-1_26} and \textsc{VRSBench} \cite{li2024vrsbenchversatilevisionlanguagebenchmark}. In this pipeline, generic VLMs generate initial drafts guided by coarse, noisy metadata-spatially and temporally aligned information such as land-cover maps and crowdsourced OSM tags. Successive VLM-assisted refinement steps polish these drafts, and human reviewers remove any text that is visually ungrounded or temporally mismatched. By pairing multi-satellite, multi-decadal Landsat scenes with reliable language supervision, \LandsatAU{} provides the first solid foundation for training and evaluating VLMs aimed at affordable, long-term Earth monitoring.

Our findings highlight a substantial gap between the capabilities of generic VLMs and the demands of long-term, low-resolution satellite imagery. Off-the-shelf VLMs perform poorly on Landsat-style data. For instance, the open-source VLM EarthDial achieves a captioning score of \textbf{0.07 SPIDEr} and an overall VQA accuracy of \textbf{0.48}, with particularly low scores of \textbf{0.23} on Agro-Phenology Reasoning and \textbf{0.10} on Cloud-Occlusion Assessment. However, after lightweight fine-tuning of the Qwen2.5-VL-7B model on our \LandsatAU{} dataset, performance improves significantly, with captioning scores rising from \textbf{0.11} to \textbf{0.31 SPIDEr} and VQA accuracy increasing from \textbf{0.74} to \textbf{0.87}. These results suggest that scalable, cost-effective Earth monitoring with VLMs is feasible, but only when using data that captures Landsat’s unique resolution, sensor diversity, and temporal depth.

The main contributions of our work are as follows:
\begin{itemize}
    \item \textbf{\LandsatAU{} dataset.} A large-scale, open-source vision-language dataset for the Landsat program featuring \textbf{30m resolution images}. It includes $196,262$ image-caption pairs and $17,725$ human-verified VQA samples, covering \textbf{four Landsat missions} from \textbf{1988 to 2024}.
    \item \textbf{Bootstrapped curation pipeline.} A semi-automatic data generation framework that leverages spatially and temporally aligned but noisy metadata (e.g., land-cover maps, OSM tags), generic VLM prompting, iterative refinement, and human verification to produce high-quality captioning and VQA annotations.
\item \textbf{Comprehensive evaluation.} Benchmarks on eight generic VLMs reveal substantial limitations in both captioning and VQA, especially in spatial reasoning and counting, while fine-tuning on \LandsatAU{} leads to significant improvements across tasks.

\end{itemize}

\section{Related Work}

\subsection{Generic Vision-Language Datasets}

Large-scale image-text datasets play an important role in the development of VLMs. Pioneering VLM datasets like \textsc{Flickr30k} \cite{DBLP:journals/corr/PlummerWCCHL15} and \textsc{MS COCO} \cite{DBLP:journals/corr/LinMBHPRDZ14} relied on costly human annotation. The \textsc{SBU Captioned Photo Dataset} \cite{NIPS2011_5dd9db5e} and \textsc{Conceptual Captions 3M} \cite{sharma2018conceptual} expanded the scale of VLM datasets to several million image-text pairs by using web images and their associated alt-text. Using a similar approach and adding quality control from machine learning models, VLM datasets like \textsc{Conceptual 12M} \cite{DBLP:journals/corr/abs-2102-08981}, \textsc{LAION-5B} \cite{schuhmann2022laion5bopenlargescaledataset}, and \textsc{COYO-700M} \cite{kakaobrain2022coyo-700m} further increased the scale to hundreds of millions or even billions of pairs. The success of models like CLIP \cite{DBLP:journals/corr/abs-2103-00020} and ALIGN \cite{DBLP:journals/corr/abs-2102-05918} demonstrated that even large-scale datasets with noisy information can significantly contribute to VLM development. Many researchers are working to improve dataset quality by using advanced VLMs, like BLIP \cite{DBLP:journals/corr/abs-2201-12086} and InstructBLIP \cite{dai2023instructblipgeneralpurposevisionlanguagemodels}, to refine noisy data and generate higher-quality annotations. Furthermore, models such as LLaVA \cite{liu2023visualinstructiontuning} and MiniGPT-4 \cite{zhu2023minigpt4enhancingvisionlanguageunderstanding} generate synthetic captions to build large-scale training datasets and reduce dataset costs. 

\subsection{Remote-Sensing Vision-Language Datasets}

The evolution of remote sensing VLMs has mirrored that of the general VLM community. Datasets like \textsc{UCM-Captions} and \textsc{Sydney-Captions} \cite{DBLP:journals/corr/abs-1712-07835} consisted of only a few hundred images with domain expert labels. To increase dataset scale, \textsc{NWPU-Captions} \cite{9866055} and \textsc{RSICD} \cite{rosario2023satellitecaptioninglargelanguage} retrieved imagery and metadata from commercial map services and then used crowdsourcing to edit the captions. With imagery from open-source satellite platforms, \textsc{SkyScript} \cite{wang2023skyscriptlargesemanticallydiverse} and \textsc{OpenSentinelMap} \cite{9856983} utilized open-source tags from free map services to create captions. However, this approach introduces temporal misalignments between static map tags and dynamic landcover. Following the success of synthetic datasets in general VLMs, remote sensing projects such as \textsc{RS5M} \cite{10679571}, SkySenseGPT \cite{luo2024skysensegptfinegrainedinstructiontuning}, ChatEarthNet \cite{yuan2024chatearthnetglobalscaleimagetextdataset}, \textsc{Git-10M} \cite{10988859}, and RS-LLaVA \cite{rs16091477} have scaled to tens of millions of image-text pairs by using prompted LLMs to synthesize instructions. Meanwhile, task-specific VQA benchmarks such as RSIVQA \cite{DBLP:journals/corr/abs-2003-07333} continue to reveal VLM weaknesses in counting, spatial reasoning, and domain transfer. Despite recent progress, the historical Landsat archive remains underutilized in VLM research. For example, the recent \textsc{EarthDial} \cite{soni2025earthdialturningmultisensoryearth}, despite its multi-sensory approach, includes only imagery from Landsat 8. \LandsatAU{} addresses this gap by providing images from four Landsat sensors that span from 1988 to 2024, thereby enabling long-term, continental-scale studies with an open-source VLM dataset.

\begin{figure*}[t]  
  \centering
  \includegraphics[width=\linewidth]{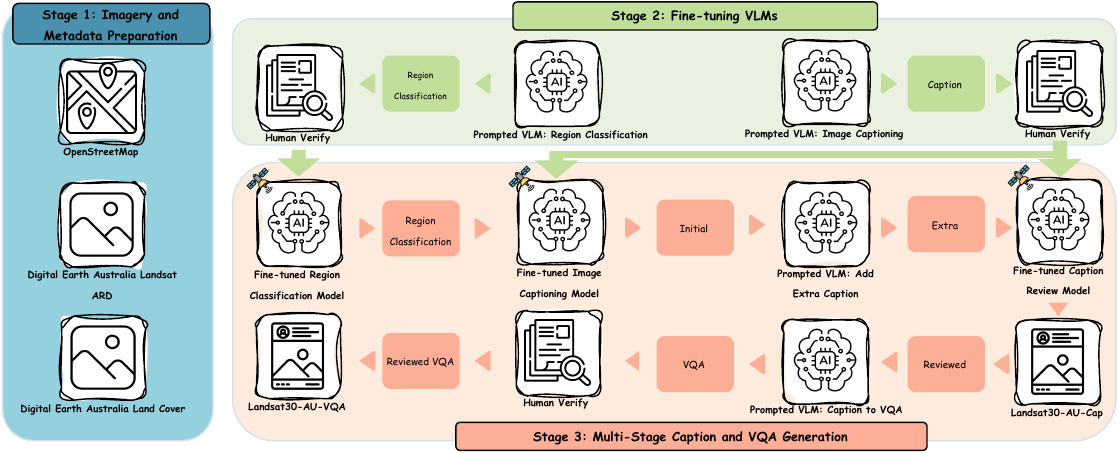}
  \caption{Overview of the \LandsatAU{} dataset construction pipeline. Stage 1: Sources Landsat imagery and collects metadata. Stage 2: Adapts VLMs into specialized modules for region classification, caption generation, and review. Stage 3: Produces large-scale annotations via iterative VLM refinement and human verification. }\label{fig:pipeline}
\end{figure*}

\begin{table*}[t]
  \centering

  \begin{subtable}[t]{\linewidth} 
    \centering
    \small
    \setlength{\tabcolsep}{4pt}
    \begin{tabular}{lccccccccc}
      \hline
      Model & Subset Acc$\uparrow$ & Jaccard$\uparrow$ & Precision$\uparrow$
            & Recall$\uparrow$ & F1$\uparrow$ & LRAP$\uparrow$ & nDCG$\uparrow$
            & 1-hamming-loss$\uparrow$ & 1-ranking-loss$\uparrow$ \\ \hline
      GPT-4o              & \textbf{0.278}$^{\star}$ & \textbf{0.630}$^{\star}$ & \textbf{0.768} & \textbf{0.722} & \textbf{0.727}$^{\star}$ & \textbf{0.826}$^{\star}$ & \textbf{0.917} & \textbf{0.783}$^{\star}$ & \textbf{0.705} \\
      Qwen             & 0.220 & 0.609 & 0.735 & \textbf{0.743}$^{\star}$ & 0.720 & 0.818 & 0.912 & 0.762 & \textbf{0.710}$^{\star}$ \\
      GPT-4o w/o ft       & \textbf{0.262} & \textbf{0.612} & \textbf{0.805}$^{\star}$ & 0.676 & 0.715 & \textbf{0.816} & \textbf{0.919}$^{\star}$ & \textbf{0.776} & 0.667 \\
      Qwen w/o ft      & 0.099 & 0.450 & 0.585 & 0.588 & 0.563 & 0.708 & 0.834 & 0.653 & 0.539 \\ 
      \hline
    \end{tabular}
    \caption{Multi-label region classification metrics on the fine-tune set.}
    \label{tab:pipeline-clf_metrics}
  \end{subtable}

  \begin{subtable}[t]{\linewidth}
    \centering
    \small
    \setlength{\tabcolsep}{4pt}
    \begin{tabular}{lcccccc}
      \hline
      Model & BLEU-4$\uparrow$ & SPIDEr$\uparrow$ & BERT-F1$\uparrow$
            & 1-CHAIR-s$\uparrow$ & 1-CHAIR-i$\uparrow$ & Avg. Cap. Len. \\ \hline
      GPT-4.1 w/o ft (Initial)  & 0.160 & 0.440 & 0.902 & 0.438 & 0.843 & 149 \\
      GPT-4.1 w/o ft (Extra)    & 0.163 & 0.440 & 0.896 & 0.423 & 0.837 & 206 \\
      GPT-4.1 w/o ft (Reviewed) & 0.152 & 0.438 & 0.901 & \textbf{0.522}$^{\star}$ & \textbf{0.864}$^{\star}$ & 140 \\
      GPT-4.1 (Initial)         & \textbf{0.188}$^{\star}$ & \textbf{0.510} & \textbf{0.905}$^{\star}$ & 0.428 & 0.841 & 161 \\
      GPT-4.1 (Extra)           & 0.173 & \textbf{0.510} & 0.897 & 0.358 & 0.828 & 217 \\
      GPT-4.1 (Reviewed)        & \textbf{0.184} & \textbf{0.517}$^{\star}$ & \textbf{0.903} & \textbf{0.473} & \textbf{0.853} & 161 \\ \hline
    \end{tabular}
    \caption{Captioning metrics on the fine-tune set.}
    \label{tab:pipeline-caption_metrics}
  \end{subtable}

  \caption{Evaluation of model performance on the fine-tuning set, comparing models before (w/o ft) and after fine-tuning. The best score in each metric is marked with a star ($^{\star}$) and the top two scores are in \textbf{bold}.}
  \label{tab:pipeline-development-ft}
  
\end{table*}

\section{Dataset Construction}
\label{data-construction-pipeline}

To tackle the challenges posed by low spatial resolution, sensor diversity, and noisy metadata, we implement a three-stage, human-in-the-loop pipeline (Fig. \ref{fig:pipeline}) that steadily produces reliable, resolution-aware textual annotations for Landsat imagery. The stages are: (1) preparing imagery and auxiliary metadata, (2) fine-tuning generic VLMs on Landsat-specific tasks, and (3) generating captions and VQA items through multi-stage refinement.

\subsection{Stage 1: Imagery and Metadata Preparation}

\paragraph{Landsat imagery.}  
We source atmospherically and geometrically corrected imagery from the Digital Earth Australia (DEA) Analysis Ready Data (ARD) archive \cite{ga_analysis_ready_data}. We use Bands 4 (Red), 3 (Green), and 2 (Blue) to generate over 400,000 $256 \times 256$ pixel RGB tiles at 30-meter GSD.

\paragraph{OpenStreetMap tags.}  
OpenStreetMap (OSM) is a crowdsourced geospatial database containing fine-grained vector annotations such as \texttt{clinic}, \texttt{road}, and \texttt{footpath}. For each tile, we extract OSM tags located within its footprint and map them to coarser, Landsat-visible categories using a predefined lookup table (e.g., \texttt{clinic} $\rightarrow$ \texttt{urban fabric}). These tags provide supplemental semantic cues when the associated objects are large enough to be resolved at 30 m GSD.

\paragraph{Land cover reference.}  
The DEA Land Cover product \cite{ga_dea_landcover_v2} provides annually updated, pixel-level classifications (e.g., artificial surfaces, natural bare, water) derived from satellite observations. We extract the dominant land cover class for six fixed spatial regions within each image: top-left, top-right, bottom-left, bottom-right, center, and entire tile. These structured region-level labels support downstream tasks such as region classification and guided captioning.

\subsection{Stage 2: Fine-tuning VLMs for Landsat Tasks}

Generic VLMs are not calibrated for 30 m imagery or Landsat’s mission-specific colour shifts.  
We therefore divide the adaptation process into three lightweight modules: \emph{region classification}, \emph{caption generation}, and \emph{caption review}, and fine-tune each using a small, manually verified subset.

\paragraph{Region classification.} 
Following ChatEarthNet~\cite{yuan2024chatearthnetglobalscaleimagetextdataset}, each $256{\times}256$ tile is partitioned into six zones: top-left, top-right, bottom-left, bottom-right, center, and entire tile.  
For each zone, we assign the dominant land-cover class based on the DEA annual land-cover raster \cite{ga_dea_landcover_v2}, using a taxonomy of coarse land-cover types (e.g., cropland, forest, water, urban).  

We manually validate 2,722 such tile-region label sets and fine-tune GPT-4o (gpt-4o-2024-08-06) from OpenAI \cite{openai2024gpt4ocard} on this task. For correctness, we use Subset Accuracy \cite{10.1007/978-3-540-24775-3_5}. For set similarity and per-label quality, we employ the Jaccard Index, Precision, Recall, and F1-score. Ranking performance is measured with Label-Ranking Average Precision (LRAP) \cite{10.5555/2980539.2980628} and nDCG \cite{jarvelin2002cumulated}. To incorporate error rates, we report (1 - Hamming Loss) and (1 - Ranking Loss), ensuring higher values are consistently better across all metrics. The fine-tuned model achieves Subset Accuracy 0.28 and Jaccard 0.63, outperforming a Qwen2.5-VL-7B (Qwen) \cite{bai2025qwen25vltechnicalreport} baseline (Table~\ref{tab:pipeline-clf_metrics}).  

\begin{figure*}[t]
  \centering

  \begin{subfigure}[t]{0.48\textwidth}
    \centering
    \includegraphics[width=\linewidth]{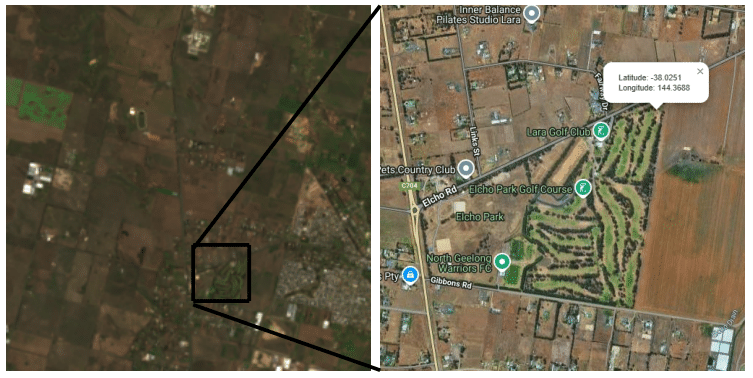}
    \caption{A golf course appears in the image. Decision: Keep.}
    \label{fig:cap-vqa-human-verify-a}
  \end{subfigure}\hfill
  \begin{subfigure}[t]{0.48\textwidth}
    \centering
    \includegraphics[width=\linewidth]{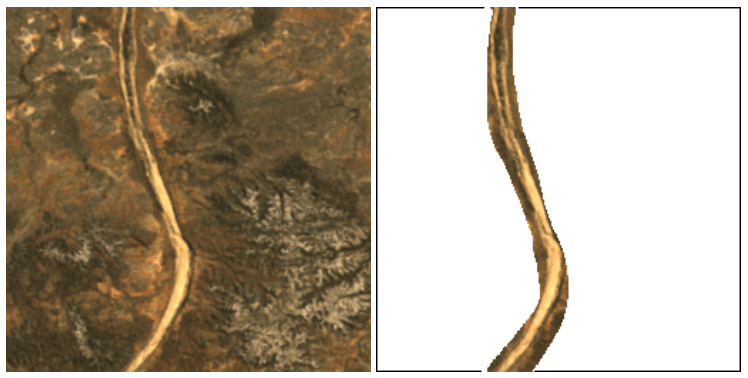}
    \caption{Which objects in the image? Fix: from highway to river.}
    \label{fig:cap-vqa-human-verify-b}
  \end{subfigure}

  \caption{Examples of the human verification process. (a) A correct caption is kept. (b) An incorrect answer is fixed.}
  \label{fig:cap-vqa-human-verify}
\end{figure*}

\paragraph{Image captioning.}  
We curated 1,005 image-caption pairs whose text explicitly referenced objects visible at 30 m and aligned with the corresponding acquisition date. All generated image-caption pairs underwent manual review. As shown in Fig. \ref{fig:cap-vqa-human-verify-a}, we used free high-resolution mapping services to verify the presence of key objects. The caption identifying a \texttt{golf course} was retained because its presence was confirmed during verification.

We fine-tuned GPT-4.1 (gpt-4.1-2025-04-14) from OpenAI \cite{openai_gpt41} on this seed dataset, resulting in captions with broader semantic coverage and improved factual grounding.  
Quantitatively, SPIDEr increased from 0.44 to 0.52, indicating better alignment with reference semantics, while 1-CHAIR-s rose from 0.44 to 0.47, reflecting fewer hallucinated object mentions.  
The average caption length also increased from 149 to 161 tokens, suggesting greater descriptive depth (Table~\ref{tab:pipeline-caption_metrics}).

\paragraph{Caption review.} 
From our initial manual review pass, we collect 9,440 image-caption  labelled \emph{keep} or \emph{delete}.  
Qwen2.5-VL-7B is fine-tuned for three epochs on these labels and thereafter prunes any sentence that is visually unsupported or temporally inconsistent, providing an automated hallucination filter for Stage 3.

Together, these three fine-tuned components supply region structure, domain-specific captioning, and scalable quality control, forming the backbone of the multi-stage caption and VQA generation pipeline.

\label{image-caption-and-caption-ft-dataset}

\begin{table*}[t]
  \centering
  \small           

  \setlength{\tabcolsep}{4pt}     
  \begin{tabularx}{\linewidth}{@{}l c l X@{}}
    \toprule
    \textbf{Type} & \textbf{\# VQA} & \textbf{Task Focus} & \textbf{Example (Fig.\,ref)} \\ 
    \midrule
    APR & 2,102 & Crop-season inference from field texture & 
      \emph{Fig.\,\ref{apr-image}}: “Wet or dry season?” Options: wet\_season, \textbf{dry\_season} \\

    COA & 2,129 & Cloud/haze usability assessment & 
      \emph{Fig.\,\ref{coa-image}}: “Scene usable despite cloud?” Options: \textbf{Fully usable}, Not usable \\

    DLC & 2,479 & Dominant land-cover type & 
      \emph{Fig.\,\ref{dlc-image}}: “Main cover type?” Options: Urban, Forest, \textbf{Field} \\

    FOD & 2,000 & Detectability of thin man-made objects & 
      \emph{Fig.\,\ref{fod-image}}: “Prominent thin structure?” Options: Railway, Pipeline, \textbf{None} \\

    MOP & 2,418 & Presence of macro-objects & 
      \emph{Fig.\,\ref{mop-image}}: “Which object is visible?” Options: Railway, Large building, \textbf{River} \\

    NUM & 2,244 & Numerosity estimation & 
      \emph{Fig.\,\ref{num-image}}: “Water-body count?” Options: Four, Two, Three, \textbf{Zero} \\

    SRI & 2,419 & Spatial-relation inference & 
      \emph{Fig.\,\ref{sri-image}}: “River vs.\ urban fabric?” Options: Only south, \textbf{Both sides}, Only north \\

    USR & 1,934 & Urban-scale recognition & 
      \emph{Fig.\,\ref{usr-image}}: “Settlement type?” Options: Major city, \textbf{Small town}, Rural \\
    \bottomrule
  \end{tabularx}
\caption{The \textsc{Landsat30-AU-VQA} taxonomy. This table outlines the eight question categories, their respective task focus, and an example for each. The correct answer in the examples is shown in \textbf{bold}.}
  \label{tab:vqa_taxonomy}

\end{table*}

\begin{figure}[tb]
  \centering
  
  \begin{subfigure}[t]{0.24\linewidth}
    \includegraphics[width=\linewidth]{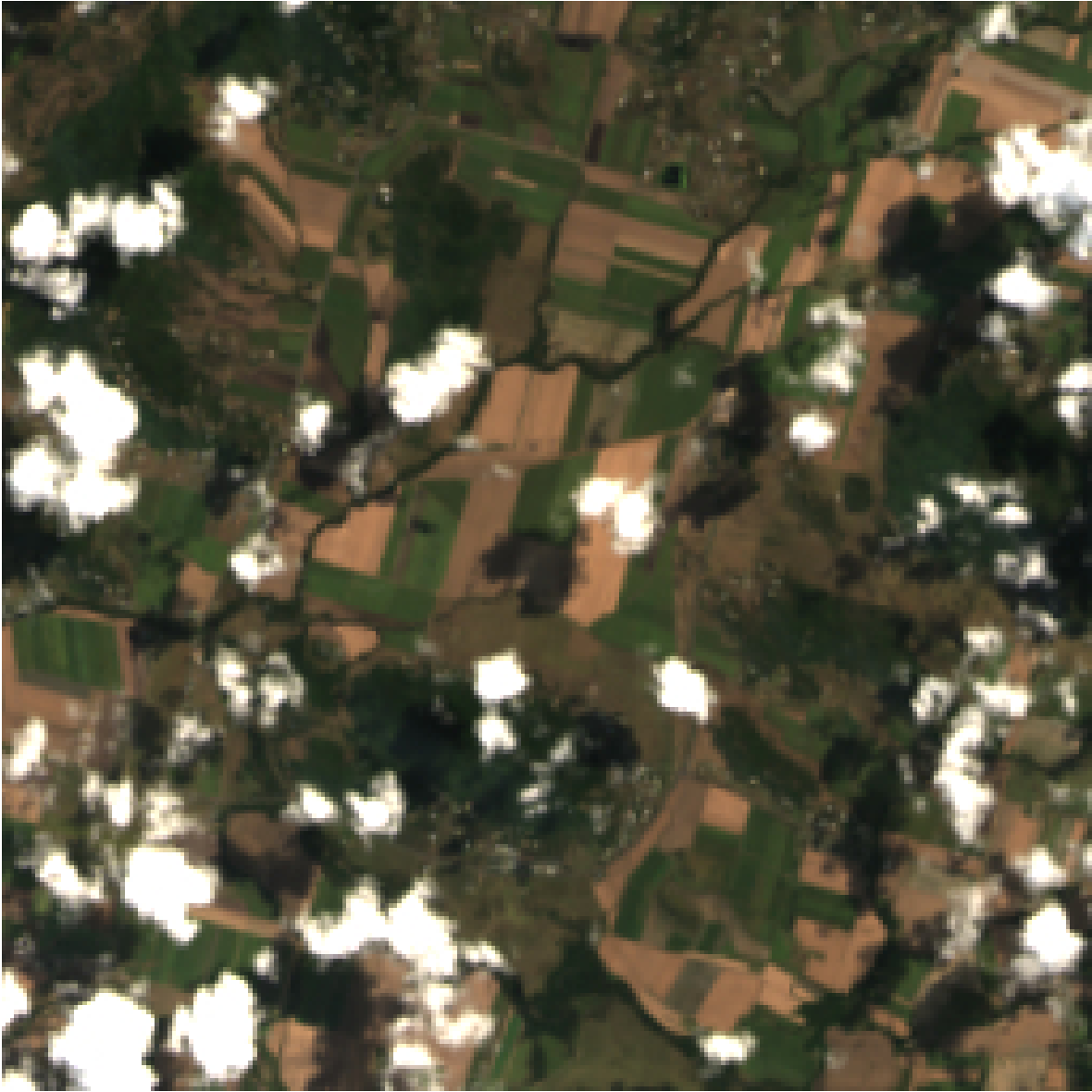}
    \caption{APR}
    \label{apr-image}
  \end{subfigure}\hfill
  \begin{subfigure}[t]{0.24\linewidth}
    \includegraphics[width=\linewidth]{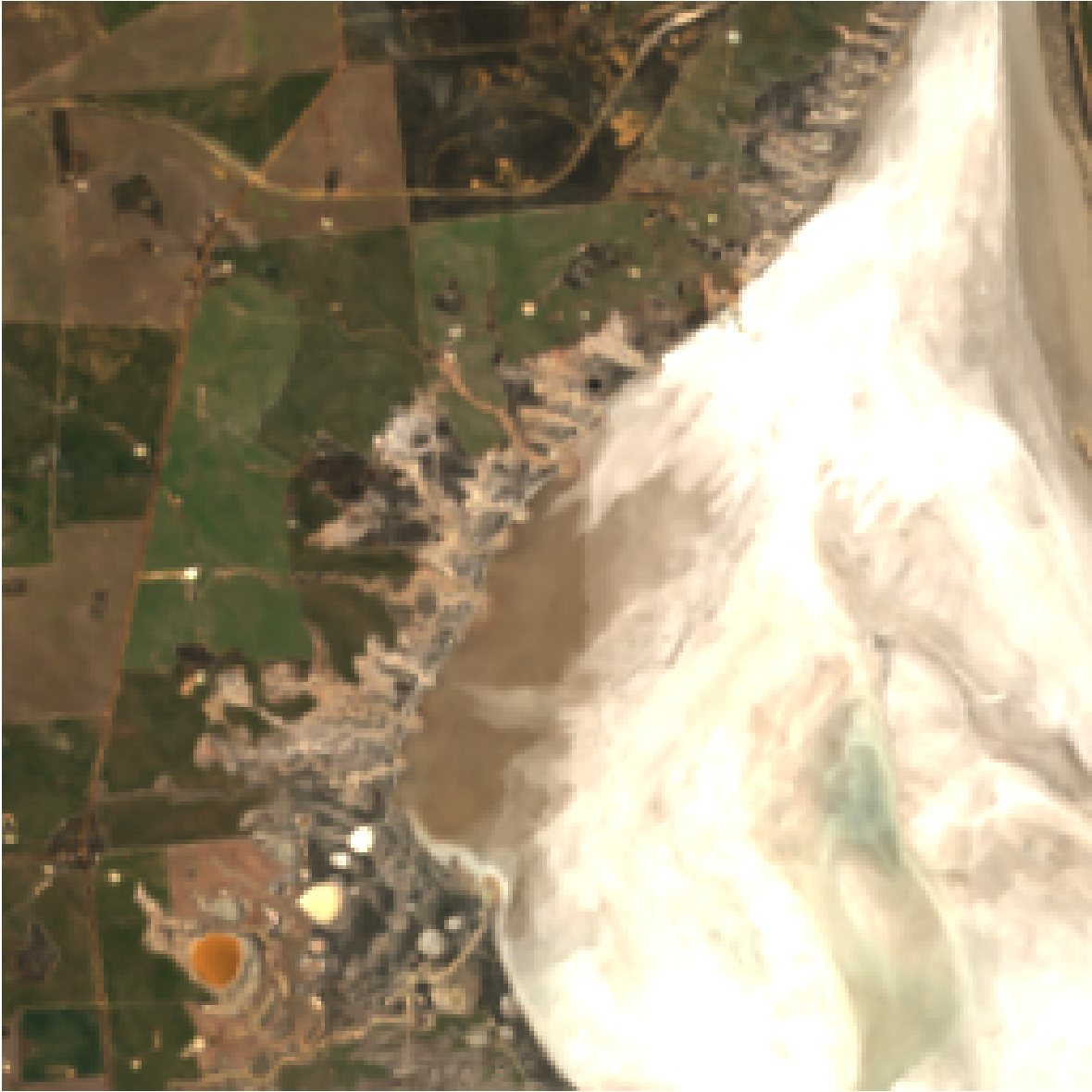}
    \caption{COA}
    \label{coa-image}
  \end{subfigure}\hfill
  \begin{subfigure}[t]{0.24\linewidth}
    \includegraphics[width=\linewidth]{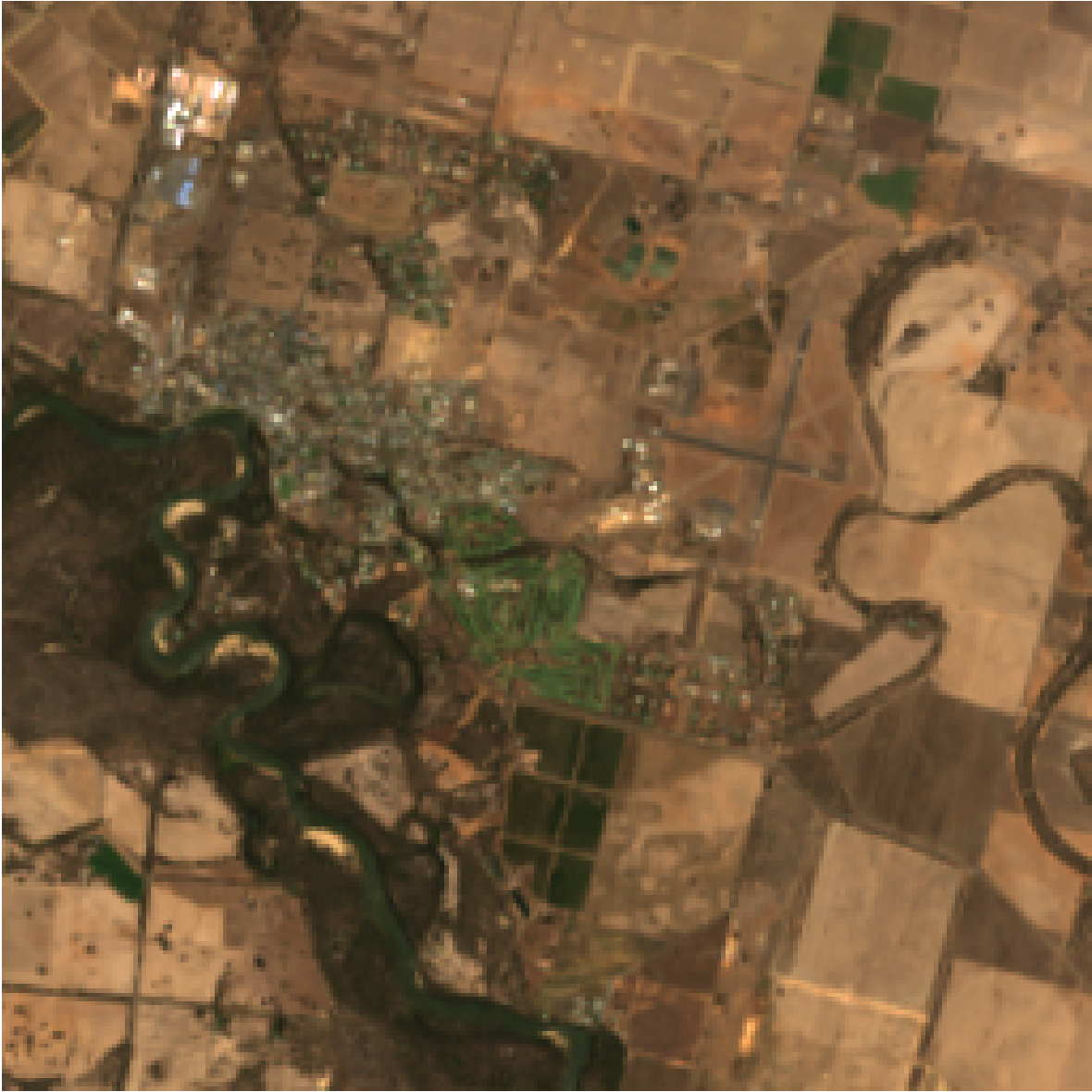}
    \caption{DLC}
    \label{dlc-image}
  \end{subfigure}\hfill
  \begin{subfigure}[t]{0.24\linewidth}
    \includegraphics[width=\linewidth]{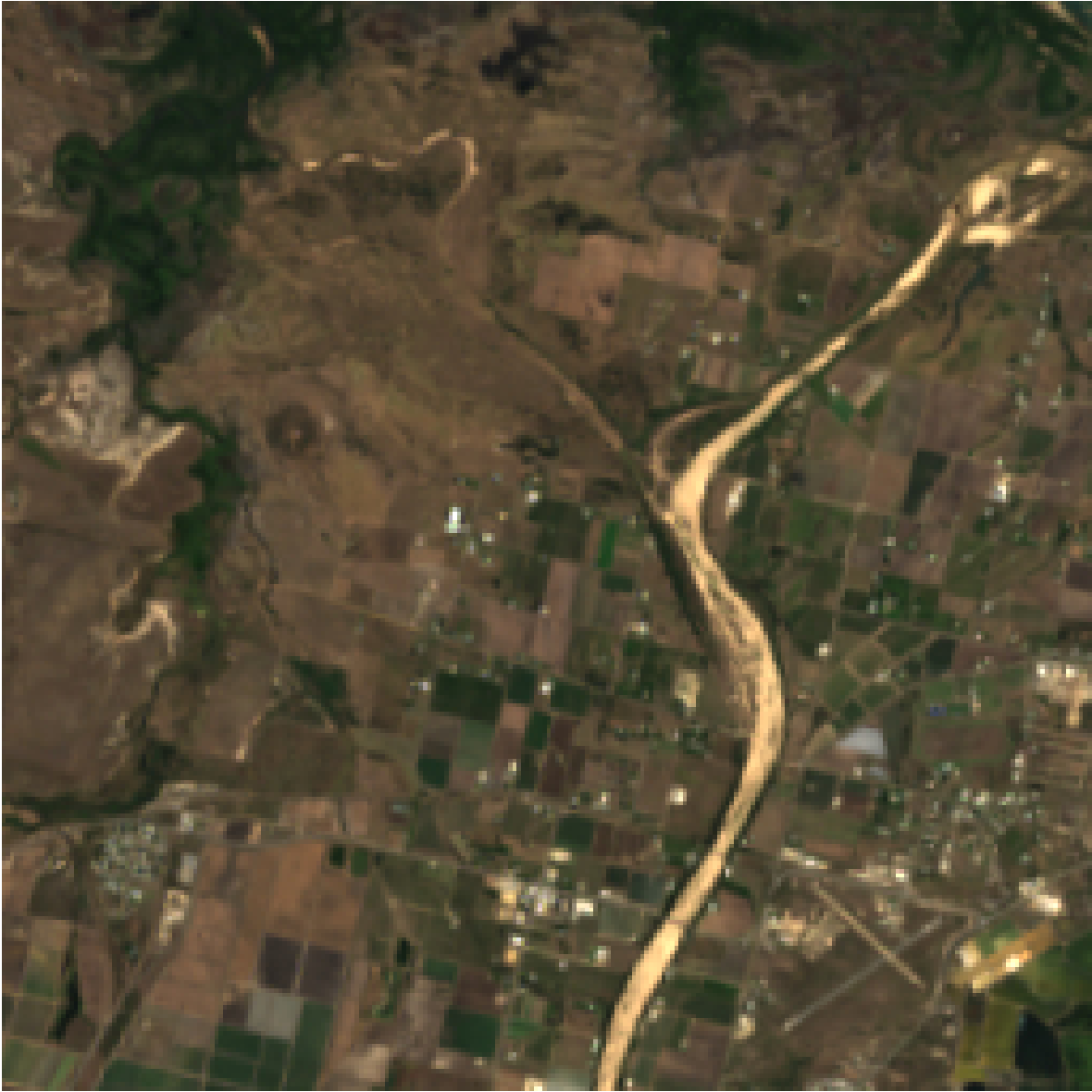}
    \caption{FOD}
    \label{fod-image}
  \end{subfigure}

  \begin{subfigure}[t]{0.24\linewidth}
    \includegraphics[width=\linewidth]{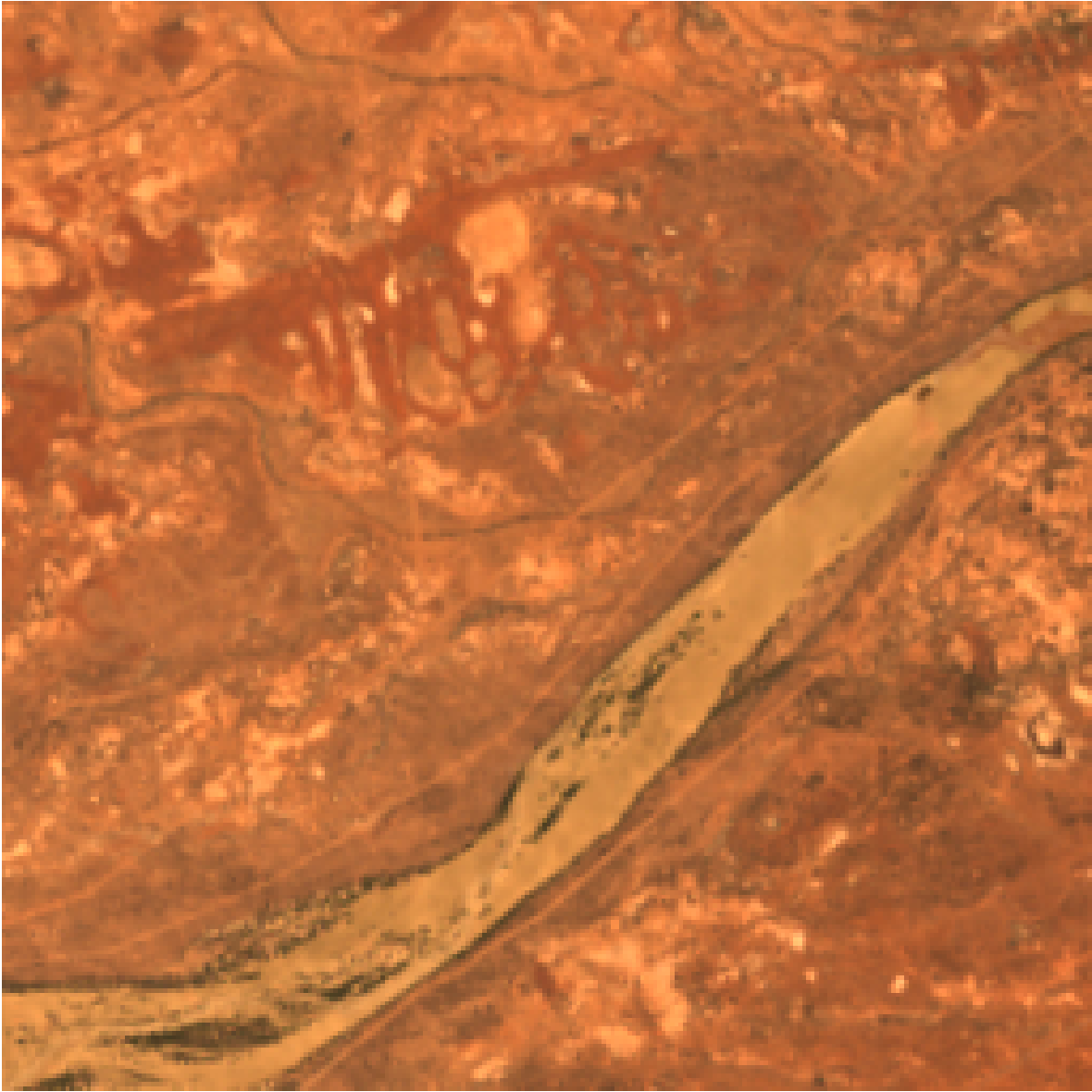}
    \caption{MOP}
    \label{mop-image}
  \end{subfigure}\hfill
  \begin{subfigure}[t]{0.24\linewidth}
    \includegraphics[width=\linewidth]{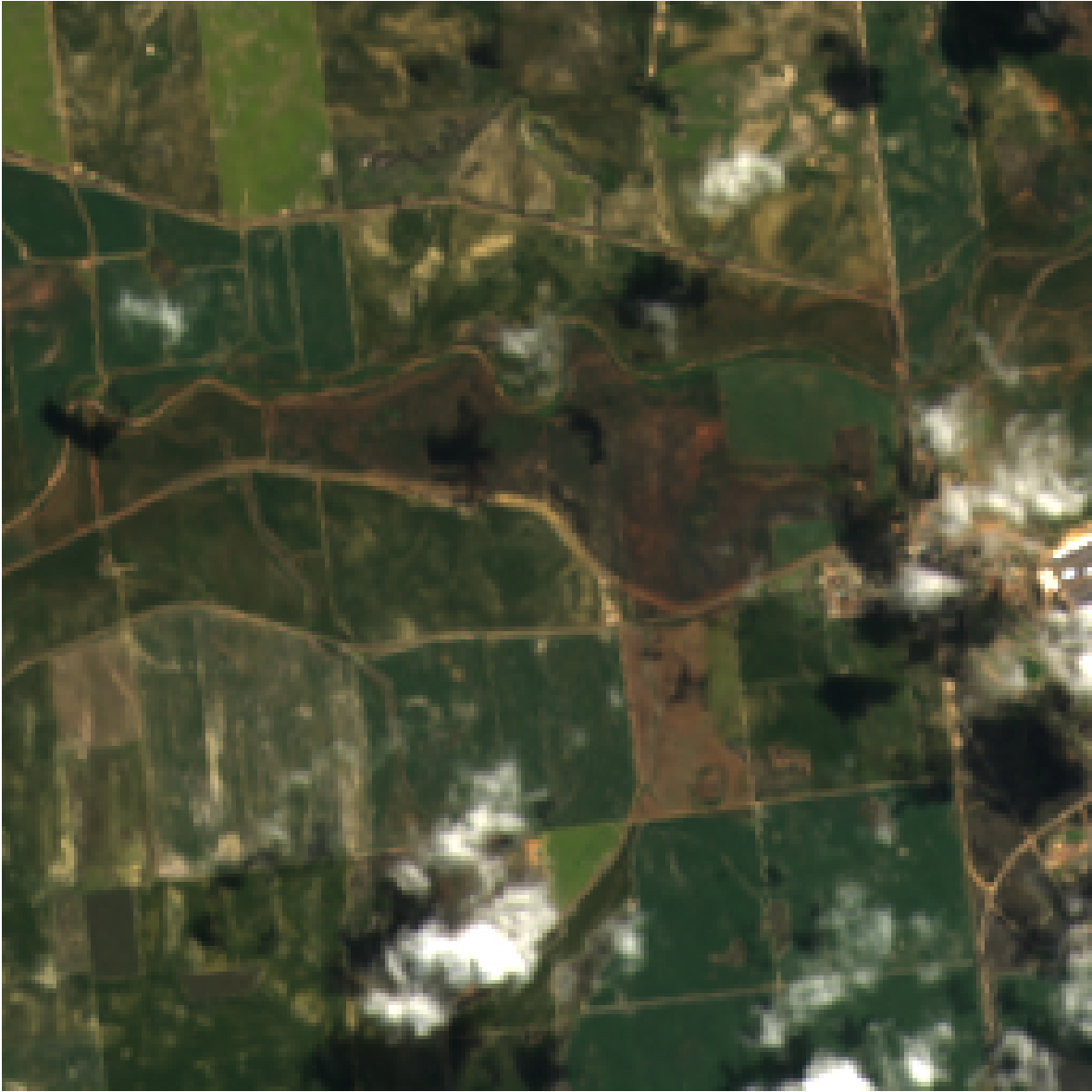}
    \caption{NUM}
    \label{num-image}
  \end{subfigure}\hfill
  \begin{subfigure}[t]{0.24\linewidth}
    \includegraphics[width=\linewidth]{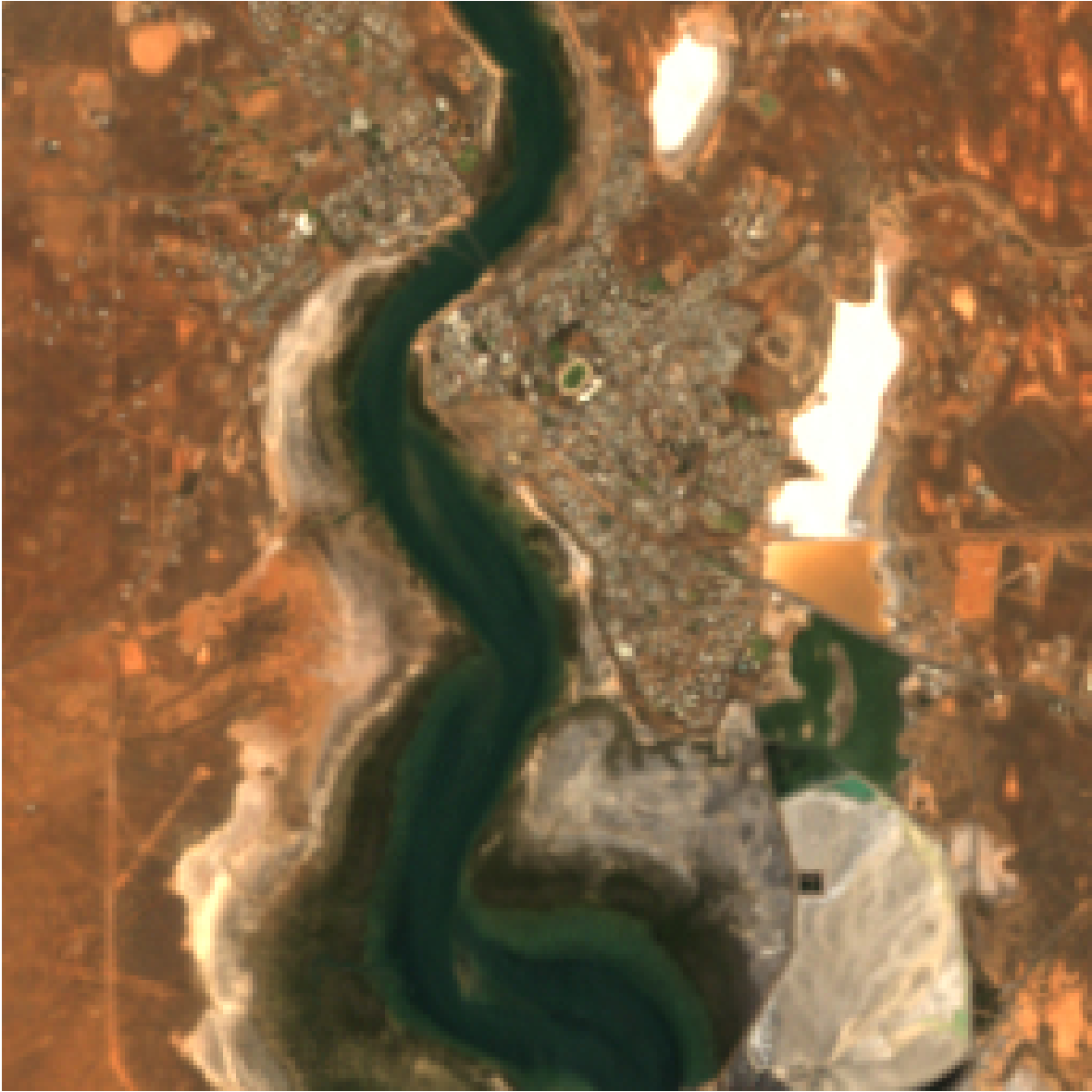}
    \caption{SRI}
    \label{sri-image}
  \end{subfigure}\hfill
  \begin{subfigure}[t]{0.24\linewidth}
    \includegraphics[width=\linewidth]{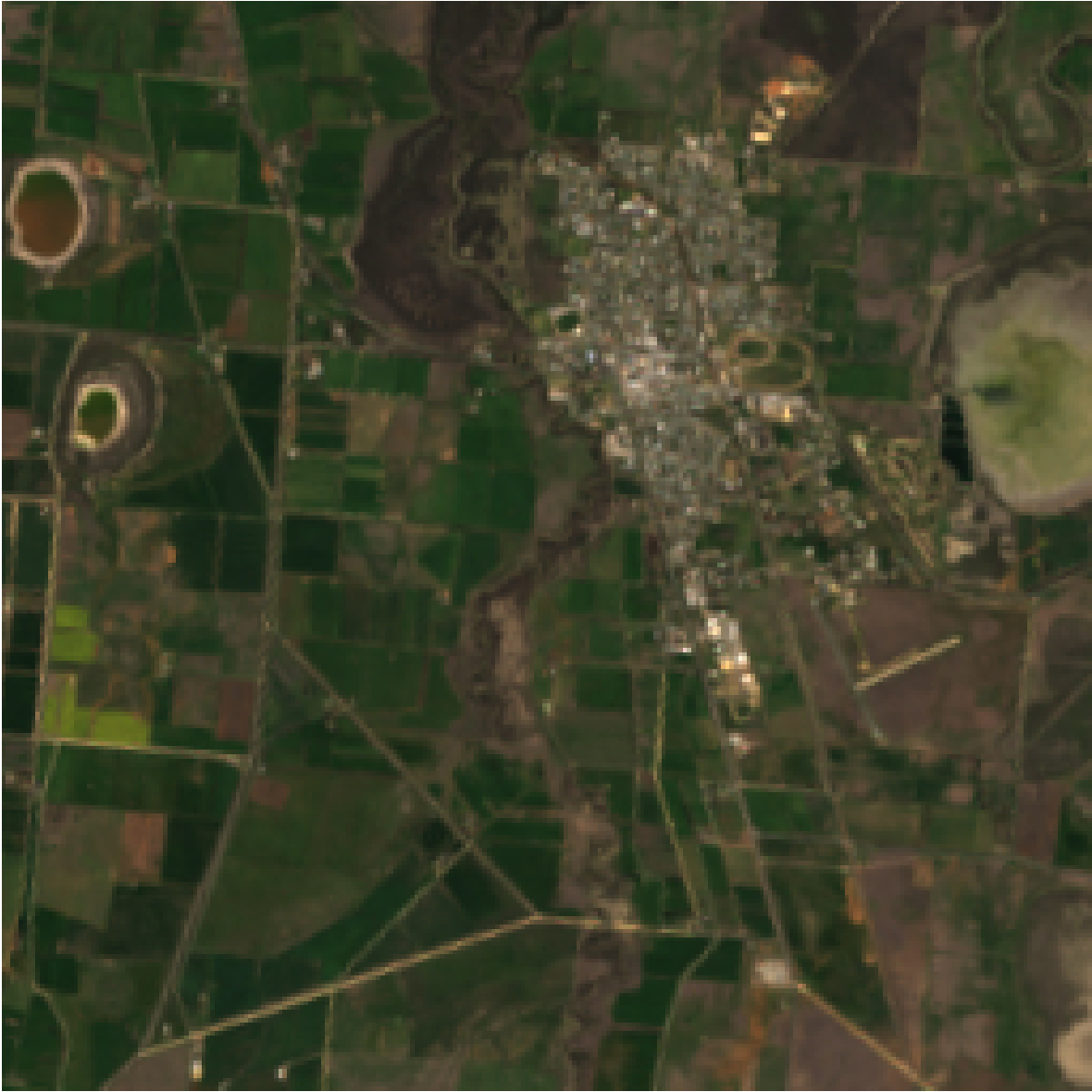}
    \caption{USR}
    \label{usr-image}
  \end{subfigure}

  \caption{Landsat30-AU-VQA categories. Representative 30-meter Landsat imagery illustrating the visual characteristics of each of the eight VQA domains.}
  \label{fig:vqa_examples}
\end{figure}

\subsection{Stage 3: Multi-Stage Caption and VQA Generation}
\label{sec:stage3}

Stage 3 uses the fine-tuned modules from Stage 2 to produce large-scale, quality-controlled annotations (Fig.~\ref{fig:pipeline}).
It involves two tasks: caption refinement and VQA generation. The caption refinement task combines model-generated drafts with VLM-based verification to ensure resolution-awareness and factual consistency, while the VQA generation task incorporates human verification to ensure answer accuracy and to increase the difficulty and diversity of the questions and options.

\paragraph{Caption refinement.}  
For each image tile, the captioning model (fine-tuned GPT-4.1) first generates an \emph{Initial} caption conditioned on region labels, OSM tags, and the image.
We then prompt Qwen2.5-VL-7B to augment the caption with missing objects and spatial relations, resulting in an \emph{Extra} version.  
Next, the caption reviewer module prunes hallucinated or temporally inconsistent content, producing the final \emph{Reviewed} caption.
To evaluate the impact of each stage, we score all three versions (\emph{Initial}, \emph{Extra}, and \emph{Reviewed}) on a held-out reference set using BLEU-4 \cite{10.3115/1073083.1073135}, SPIDEr \cite{DBLP:journals/corr/LiuZYG016}, and BERTScore-F1 \cite{DBLP:journals/corr/abs-1904-09675} for semantic quality, and CHAIR-s/i \cite{DBLP:journals/corr/abs-1809-02156} for hallucination.
As shown in Table~\ref{tab:pipeline-caption_metrics}, the \emph{Reviewed} captions provide the best overall balance (SPIDEr 0.517; 1-CHAIR-i 0.853), and are used throughout the dataset.  
This process yields 196,262 high-quality captions, which make up the \textsc{Landsat30-AU-Cap} dataset.

\paragraph{VQA generation.}  
To construct \textsc{Landsat30-AU-VQA}, we prompt GPT-4.1 to generate multiple-choice questions (MCQs) from 9,735 captioned images.  
Each MCQ is designed to assess one of eight Landsat-specific reasoning tasks (see Table~\ref{tab:vqa_taxonomy} and Fig.~\ref{fig:vqa_examples}).  
Human reviewers then refine the questions by correcting ambiguous phrasing, replacing weak distractors, and discarding low-quality items. As shown in Fig.~\ref{fig:cap-vqa-human-verify-b}, original VQA from GPT-4.1 confuses the width of a linear feature and incorrectly classified it as a highway.
We intentionally included such errors as incorrect options to force finer distinctions. Example question-answer pairs are shown in Table~\ref{tab:vqa_taxonomy} and Fig.~\ref{fig:vqa_examples}
This results in 17,725 validated question-answer pairs.

Together, these two processes complete the \LandsatAU{} corpus, providing resolution-aware, multi-sensor, and temporally grounded textual supervision for training and evaluating VLMs on real-world satellite imagery.

\section{Landsat30-AU Dataset}


This section details the two \LandsatAU{} sub-datasets, providing key statistics and a comparison with existing remote sensing vision-language corpora.

\paragraph{Landsat30-AU-Cap.}
\textsc{Landsat30-AU-Cap} consists of \textbf{196,262} image-caption pairs aligned with \textbf{low-resolution} Landsat imagery from \textbf{four satellites} spanning 36 years (\textbf{1988-2024}).  
Each caption is visually grounded and tailored to Landsat’s spatial resolution, offering detailed semantic content that reflects the constraints and opportunities of low-resolution Earth observation.  
This dataset supports training and evaluation of captioning models on real-world, multi-sensor, multi-temporal satellite imagery.

\paragraph{Landsat30-AU-VQA.}
\textsc{Landsat30-AU-VQA} contains \textbf{17,725} multiple-choice question-answer pairs covering \textbf{eight} remote sensing tasks designed to capture common reasoning challenges in low-resolution imagery.  
These include: (1) inferring cropping season from field texture (Agro-Phenology Reasoning, APR), (2) evaluating cloud and haze interference (Cloud-Occlusion Assessment, COA), (3) identifying dominant land-cover types (Dominant Land Cover, DLC), (4) detecting thin or sub-pixel structures (Fine-Object Detectability, FOD), (5) identifying large visible features (Macro-Object Presence, MOP), (6) estimating object counts (Numerosity, NUM), (7) reasoning about spatial layout (Spatial-Relation Inference, SRI), and (8) classifying settlement scale (Urban-Scale Recognition, USR).  
Examples are in Table~\ref{tab:vqa_taxonomy} and Fig.~\ref{fig:vqa_examples}.

\subsection{Comparison with Remote-Sensing VLM Datasets}

We compare \LandsatAU{} to six key remote sensing vision-language datasets: \textsc{RSICD} \cite{rosario2023satellitecaptioninglargelanguage}, \textsc{SkyScript} \cite{wang2023skyscriptlargesemanticallydiverse}, \textsc{ChatEarthNet} \cite{yuan2024chatearthnetglobalscaleimagetextdataset}, \textsc{Git-10M} \cite{10988859}, \textsc{GAIA} \cite{zavras2025gaiaglobalmultimodalmultiscale}, and \textsc{EarthDial} \cite{soni2025earthdialturningmultisensoryearth}.  

\begin{table}[t]
  \centering
  \small
  \setlength{\tabcolsep}{3pt}
  \begin{tabular}{@{}lcrcc@{}}
    \toprule
    \textbf{Dataset} & \textbf{\# img/LS img} & \textbf{\# LS Sats.} & \textbf{Geo-loc.} & \textbf{Span} \\
    \midrule
    RSICD             & 10k/0       & 0     & No  & -   \\
    SkyScript         & 5M/15k    & 2     & Yes & 2013-2023 \\
    ChatEarthNet      & 173k/0       & 0     & No  & -   \\
    Git-10M           & 16M/0    & 0     & Yes & - \\
    GAIA              & 41k/2k    & 2     & Yes & 2013-2024 \\
    EarthDial         & 11M/1.6M   & 1     & No  & 2013-2024 \\
    \midrule
    \textbf{Landsat30-AU} & \textbf{196k/196k} & \textbf{4} & \textbf{Yes} & \textbf{1988-2024} \\
    \bottomrule
  \end{tabular}
  \caption{\LandsatAU{} vs. other remote sensing VLM datasets. \emph{Span} is blank when no Landsat imagery is present.}

  \label{tab:rs_vlm_datasets}
\end{table}

\paragraph{Scope and diversity.}

Table~\ref{tab:rs_vlm_datasets} compares datasets across five key metrics: total images (\# img), the number of Landsat images (\# LS image) and number of source Landsat satellites (\# LS Sats), whether the imagery is georeferenced (Geo-loc.), and the Landsat imagery temporal span (Span), highlighting differences in scale, Landsat imagery diversity, and spatio-temporal coverage.

While EarthDial offers a larger number of Landsat images (1.6 million), it is restricted to a single satellite (Landsat 8) and lacks geolocation metadata.  
In contrast, \LandsatAU{} spans four Landsat satellites (Landsat 5, 7, 8, and 9) over a 36-year period (1988-2024), with each image accompanied by precise geographic coordinates and acquisition dates.  
This rich spatiotemporal coverage enables models to learn from diverse sensor characteristics and location-aware patterns, making \LandsatAU{} uniquely suited for multi-sensor, long-term Earth observation tasks.

\paragraph{Linguistic and semantic richness.}
Table~\ref{tab:ds_nlp_metrics} presents a comparison of caption length and lexical diversity across Landsat-related datasets.  
EarthDial does not include captions, and SkyScript provides only very short ones, averaging 9.3 words.  
GAIA offers high-quality captions, with an average length of 183.3 words and strong lexical diversity as measured by the Mean Segmental Type-Token Ratio (MSTTR) at 0.84. However, it includes only around 2,000 image-caption pairs.  
In contrast, \LandsatAU{} provides \textbf{196,262} captions with both scale and linguistic richness, featuring an average length of 165.4 words and 0.82 MSTTR.

\begin{table}[h!]
\centering
\small
\setlength{\tabcolsep}{3pt}
\begin{tabular}{@{} lccccc@ {}}
\toprule
\textbf{Dataset} & \textbf{LS Pairs} & \textbf{Vocab} & \textbf{Avg. Cap. Len.} & \textbf{MSTTR} $\uparrow$\\
\midrule
SkyScript & 15k & 1,049 & 9.3  & -  \\
GAIA & 2k & 2,325 & \textbf{183.3} $^{\star}$  & \textbf{0.84} $^{\star}$ \\
EarthDial & \textbf{1.6M} $^{\star}$ & \textbf{9,251} $^{\star}$ & - & -  \\
\midrule
\textbf{Landsat30-AU} & \textbf{196k} &\textbf{4,405} & \textbf{165.4}  & \textbf{0.82}  \\
\bottomrule
\end{tabular}
\caption{Linguistic properties of Landsat-related datasets. The best score in each metric is marked with a star ($^{\star}$), and the top two are in \textbf{bold}.}
\label{tab:ds_nlp_metrics}
\end{table}

\section{Benchmark Evaluation}

\paragraph{Task Settings.} \LandsatAU{} includes two distinct tasks for evaluating Landsat imagery understanding:

\begin{itemize}
    \item \textbf{Image-Captioning:} This is a generative captioning task requiring VLMs to produce detailed descriptions of Landsat images. We use a test set of $1,005$ human-verified image-caption pairs from Stage 2 image captioning and compare the VLM-generated captions against reference captions using BLEU-4, SPIDEr, BERT-F1, 1-CHAIR-s, 1-CHAIR-i, and Average Caption Length.
    \item \textbf{VQA:} A multiple-choice VQA task that evaluates a model's ability to understand Landsat imagery content, to infer information beyond the visual data, and address challenges specific to 30-meter GSD. We report \textbf{per-category accuracy} across eight VQA categories. We use a 15\% split of \textsc{Landsat30-AU-VQA} as the test set.
\end{itemize}

\begin{table*}[ht]                 
  \centering
  \small
  \begin{subtable}{\textwidth} 
    \centering
    \setlength{\tabcolsep}{2pt}
    \begin{tabular}{@{}llccccccc@{}}
    \toprule
    Type &  Model  & Size & BLEU-4$\uparrow$ & SPIDEr$\uparrow$ & BERTScore-F1$\uparrow$ & 1-CHAIR-s$\uparrow$ & 1-CHAIR-i$\uparrow$ & Avg. Cap. Len. \\
    \midrule
    \multirow[t]{4}{*}{Specialized}
     &EarthDial    & 4B  & 0.0210 & 0.0726 & 0.8379 & \textbf{0.5920} & 0.8197 & 140 \\
     &RS-LLaVA     & 7B  & 0.0975 & 0.2095 & 0.8874 & \textbf{0.5920} & 0.8119 & 139 \\
     & MiMo         & 7B  & 0.0338 & 0.0958 & 0.8601 & 0.3831 & 0.7805 & 168 \\
     & GLM-V        & 9B  & 0.0420 & 0.1177 & 0.8668 & \textbf{0.6259}$^{\star}$ & \textbf{0.8496} & 155 \\
    \midrule
    \multirow[t]{4}{*}{General}
      &Qwen         & 7B  & 0.0350 & 0.1114 & 0.8693 & 0.4697 & 0.7959 & 124 \\
     &LLaVA        & 8B  & 0.0258 & 0.1286 & 0.8643 & 0.5483 & 0.8437 & 103 \\
    & Llama        & 11B  & 0.0726 & 0.1695 & 0.8800 & 0.5483 & 0.8296 & 147 \\
     &Gemma 3      & 12B  & 0.0542 & 0.1246 & 0.8751 & 0.3572 & 0.8019 & 149 \\
    \midrule
    \multirow[t]{2}{*}{General with ft}
    &Qwen-ft      & 7B  & \textbf{0.1395}$^{\star}$ & \textbf{0.3054}$^{\star}$ & \textbf{0.8935}$^{\star}$  & 0.4657 & \textbf{0.8549}$^{\star}$ & 157 \\
    &Llama-ft     & 11B  & \textbf{0.1129} & \textbf{0.2767} & \textbf{0.8914} & 0.5224 & 0.8016 & 124 \\
    \bottomrule
    \end{tabular}
    \caption{Performance on the image captioning task.}
    \label{tab:caption_metrics}
  \end{subtable}
    
  \begin{subtable}{\textwidth} 
    \centering
    \setlength{\tabcolsep}{5pt}
    
    \begin{tabular}{@{}llccccccccccc@{}}
      \toprule
      Type & Model  & Size & APR$\uparrow$ & COA$\uparrow$ & DLC$\uparrow$ & FOD$\uparrow$ &
             MOP$\uparrow$ & NUM$\uparrow$ & SRI$\uparrow$ & USR$\uparrow$ & Overall \\
      \midrule
      \multirow[t]{4}{*}{Specialized}
     & EarthDial  & 4B & 0.2349 & 0.1034 & 0.7527 & \textbf{0.9900} & 0.6116 & 0.4362 & 0.5124 & 0.1552 & 0.4829 \\ 
     &RS-LLaVA   & 7B & \textbf{0.6857} & 0.8088 & 0.7124 & 0.8700 & 0.6309 & 0.4985 & 0.2617 & 0.1034 & 0.5724 \\
     & MiMo       & 7B & 0.4000 & 0.4577 & 0.9247 & 0.9333 & \textbf{0.8430} & \textbf{0.6142} & \textbf{0.9421}$^{\star}$ & 0.8897 & \textbf{0.7555} \\ 
     & GLM-V      & 9B & 0.4571 & 0.3636 & 0.7285 & 0.6267 & 0.6749 & 0.5863 & 0.6997 & 0.8828 & 0.6287 \\ 
    \midrule
    \multirow[t]{4}{*}{General}
     & Qwen       & 7B & 0.2984 & \textbf{0.8966} & \textbf{0.9409} & 0.7167 & 0.7603 & 0.5312 & 0.9284 & 0.8207 & 0.7428 \\
     &LLaVA      & 8B & 0.3937 & 0.7900 & 0.8306 & 0.5900 & 0.7245 & 0.4659 & 0.8512 & 0.1034 & 0.6096 \\
    &Llama      & 11B & 0.3111 & 0.8558 & 0.6022 & 0.6633 & 0.7135 & 0.5757 & 0.8953 & 0.1034 & 0.6025 \\
     &Gemma 3    & 12B & 0.6730 & 0.8150 & 0.9220 & 0.4533 & 0.7934 & 0.3234 & 0.9311 & \textbf{0.9310}$^{\star}$ & 0.7356 \\
    \midrule
    \multirow[t]{2}{*}{General with ft}
     & Qwen-ft    & 7B & \textbf{0.7016}$^{\star}$ & \textbf{0.9530}$^{\star}$ & \textbf{0.9651}$^{\star}$ & \textbf{1.0}$^{\star}$   & \textbf{0.8678}$^{\star}$ & \textbf{0.6588}$^{\star}$ & 0.9229 & \textbf{0.8966} & \textbf{0.8710}$^{\star}$ \\
     & Llama-ft   & 11B & 0.5238 & 0.8558 & 0.8682 & \textbf{1.0}$^{\star}$    & 0.8402 & 0.6024 & \textbf{0.9339} & 0.1276 & 0.7315 \\ 
    \bottomrule
    \end{tabular}
    \caption{Performance on the VQA task, reported as accuracy per category}
    \label{tab:vqa_metric}
  \end{subtable}

  \caption{Evaluation of VLMs on Landsat30-AU. Bold indicates a top-2 score. $^{\star}$ indicates the best score.}
  \label{tab:benchmark_result}
  
\end{table*}

\paragraph{Implementation Details.} To structure our evaluation, we group the models based on their training domain. The Specialized category comprises remote sensing VLMs, including EarthDial \cite{soni2025earthdialturningmultisensoryearth} and RS-LLaVA \cite{rs16091477}, and reasoning VLMs, such as GLM-4.1V (GLM-V) \cite{vteam2025glm41vthinkingversatilemultimodalreasoning} and MiMo-VL (MiMo) \cite{coreteam2025mimovltechnicalreport}. The General category consists of foundational models like Qwen2.5-VL (Qwen), Gemma 3 (Gemma3) \cite{gemmateam2025gemma3technicalreport}, Llama-3.2 (Llama) \cite{grattafiori2024llama3herdmodels}, and LLaVA-OneVision (LLaVA) \cite{li2024llavaonevisioneasyvisualtask}. We ran the two reasoning models in a zero-shot setting, enforcing a maximum output of 8,192 tokens, while the remaining models were evaluated in a one-shot setting.

Furthermore, we fine-tuned two of the general models, Qwen and Llama (creating Qwen-ft and Llama-ft), using LoRA \cite{DBLP:journals/corr/abs-2106-09685} on 15\% of the respective training data for each task (\textsc{Landsat30-AU-Cap} for captioning and \textsc{Landsat30-AU-VQA} for VQA).

\subsection{RQ1: How do Specialized VLMs perform compared to General models?}

\paragraph{Settings.} We analyze the performance of Specialized VLMs including remote sensing VLMs (EarthDial, RS-LLaVA) and reasoning VLMs (GLM-V, MiMo) against general models (without fine-tune).

\paragraph{Results.} The specialized models exhibit distinct trade-offs. RS-LLaVA proves to be a competent semantic captioner, while EarthDial lags significantly (Table~\ref{tab:caption_metrics}). Both models show strong sentence-level hallucination control, suggesting a shared cautiousness in their design. However, their VQA performance reveals critical flaws: EarthDial fails on tasks like APR, COA, NUM, SRI, and USR, while RS-LLaVA surprisingly struggles with fundamental SRI and USR, achieving the lowest score of all models.We hypothesize this stems from a domain mismatch between their training corpora and our Landsat imagery. The reasoning VLM MiMo achieves the second-highest overall VQA score ($0.7555$), notably excelling in NUM and MOP, showcasing the value of its chain-of-thought capabilities. However, its verbose captions lead to the worst sentence-level hallucination rate of $0.3831$ on the 1-CHAIR-s metric. GLM-V is the most factually grounded captioner with the best hallucination score, but its VQA performance is unremarkable. Ultimately, neither remote sensing nor reasoning VLMs demonstrate the consistent, all-around competence required for robust Landsat imagery analysis, as the stark performance divergence even within the same category reveals strong, conflicting biases inherited from their unique training domains.

\subsection{RQ2: Can fine-tuning improve VLM performance in Landsat imagery understanding?}

\paragraph{Settings.} We compare the performance of the base Qwen and Llama models against their fine-tuned ones (Qwen-ft, Llama-ft) on both the captioning and VQA tasks.

\paragraph{Results.} As shown in Table~\ref{tab:benchmark_result}, fine-tuning provides a decisive performance boost on both models. On the captioning task, Qwen-ft achieves state-of-the-art results, leading in BLEU-4 ($0.1395$), SPIDEr ($0.3054$), and BERTScore-F1 ($0.8935$), while simultaneously demonstrating strong hallucination control, with a 1-CHAIR-i score of $0.8549$. While Llama-ft also saw a substantial ~63\% gain in its SPIDEr score, it revealed a nuanced trade-off, with a slight increase in object hallucination. The most compelling evidence lies in the VQA tasks, where fine-tuning specifically improved performance on domain-specific challenges. For instance, accuracy on APR more than doubled for Qwen-ft, while both fine-tuned models achieved perfect scores on FOD, effectively learning the resolution limits of the imagery. Qwen-ft achieves the highest overall accuracy ($0.8710$) and secures top scores in six of the eight reasoning categories. These results unequivocally demonstrate that even limited, efficient fine-tuning is critical for adapting VLMs to the specific visual and logical challenges of Landsat imagery analysis.

\subsection{RQ3: What are the strengths and weaknesses of VLMs on Landsat imagery?}

\paragraph{Settings.} We analyze the per-category VQA accuracies across all VLMs in Table~\ref{tab:vqa_metric}.

\paragraph{Results.} Models consistently excel at direct perceptual tasks, such as identifying dominant land cover (DLC), confirming the presence of macro-objects (MOP), or correctly assessing the absence of sub-pixel features (FOD). This indicates a strong baseline for grounded visual recognition.

However, performance degrades significantly as tasks demand more abstract or contextual reasoning. Numerosity (NUM) emerges as a universal bottleneck across all models. Similarly, tasks requiring contextual assessment of the entire scene, such as judging urban scale (USR) or cloud usability (COA), produce highly polarized results, suggesting that only some models have learned the necessary holistic interpretation skills. The most abstract reasoning tasks, like inferring seasonality from texture (APR) or deducing complex spatial relationships (SRI), remain the most challenging and are often the primary beneficiaries of targeted fine-tuning. This pattern suggests that while current VLMs have mastered direct perception for Landsat imagery.

\section{Conclusion}


We introduce a new dataset derived from optical imagery across the Landsat 5, 7, 8, and 9 missions. This dataset is organized into two subsets: \textsc{Landsat30-AU-Cap}, containing $196,262$ image-captioning pairs, and \textsc{Landsat30-AU-VQA}, comprising $17,725$ human-verified VQA samples. Both components are built from 30-meter resolution Landsat imagery and are curated to facilitate VLM training and validation within this specific domain. Our benchmark evaluation reveals that while fine-tuning is critical for adapting models, significant challenges remain in complex tasks, thereby highlighting key areas for future VLM development.

\section*{Acknowledgements}
Contribution list:
\paragraph{Sai Ma:} Conceptualization; Data Curation; Methodology; Investigation; Formal Analysis; Writing – Original Draft; Writing – Review \& Editing. 

\paragraph{Zhuang Li:} Conceptualization; Methodology; Writing – Original Draft; Writing – Review \& Editing; Supervision. 

\paragraph{John A. Taylor:} Supervision; Conceptualization; Resources; Project Administration.

\bibliography{aaai2026}

\input{Landsat30-AU-Appendix}

\end{document}

%% file: Landsat30-AU-Appendix.tex
\appendix

\section{Dataset Construction}
\subsection{Stage 1: Imagery and Metadata Preparation}

\subsubsection{Landsat imagery.}

We sourced our Landsat imagery from the Digital Earth Australia (DEA) Analysis-Ready Data (ARD) product. From this data, we created $256 \times 256$\,px true-color RGB patches by utilizing the 30-meter resolution \texttt{nbart\_red}, \texttt{nbart\_green}, and \texttt{nbart\_blue} bands. To ensure systematic and consistent geographic coverage, we partitioned Australia into a static grid of approximately 2,500 Areas-of-Interest (AOIs), each covering a $7,680 \times 7,680$\,m\(^2\) area.

To construct a long-term dataset from 1988 to 2024, we implemented a temporal sampling strategy. For each AOI, we aimed to select one image per year. To introduce seasonal diversity, the search for a suitable image began in a different quarter each year, determined by the year number modulo 4. For instance, the search for an image in 2001 would commence from the second quarter (April 1st). The first image found that passed the initial quality filter was selected.

Our initial filtering criterion was based on the ARD-provided metadata, keeping only patches marked as at least $99.5\,\%$ cloud-free. This process generated a draft dataset of roughly $400,000$ patches, each with its geohash, capture time, and satellite metadata. However, manual inspection revealed that the ARD cloud mask was unreliable, allowing many cloudy images to pass. Furthermore, a significant portion of the patches were semantically uninteresting (e.g., homogeneous water or desert scenes).

To overcome these limitations, we implemented a second, more robust filtering stage. We prompted a Qwen2.5-VL-7B model to analyze every patch and provide a more accurate cloud cover percentage, effectively removing images with residual clouds and improving the overall quality of the dataset. The specific prompt used for this refinement is available in Listing~\ref{cloudy-or-clear-prompt}. This multi-stage approach resulted in our final \textsc{Landsat30-AU} dataset, comprising $196,262$ high-quality Landsat images.

\begin{listing}[ht]%
\caption{Cloudy Or Clear Prompt.}%
\label{cloudy-or-clear-prompt}%
\begin{lstlisting}[basicstyle=\ttfamily\small,breaklines=true,language=Python]
system_prompt = """You are an advanced assistant specializing in analyzing optical satellite images. Your task is to classify each satellite image as either "cloudy" or "clear".

Definitions:
Cloudy: The majority of the image is covered by clouds, obscuring most of the Earth's surface, OR if the image is dominated by features or artifacts (such as sensor bands, stripes, or areas with missing data) that prevent a clear view of ground features.
Clear: The image is mostly free of clouds, and the surface of the Earth is clearly visible.

Instructions:
If clouds or visual obstructions (e.g. striping, missing data, sensor artifacts, over-exposure) cover most of the image and you cannot clearly see the ground features, classify as "cloudy".
If the ground and surface features are mostly visible, classify as "clear".
Respond only with "cloudy" or "clear"."""

user_prompt = "Please classify the image as either 'cloudy' or 'clear'."
}
\end{lstlisting}
\end{listing}

\subsubsection{OpenStreetMap tags.}
\label{OSM-mapping}

To enrich our imagery with ground-level context, we sourced land use information from OpenStreetMap (OSM). For each AOI, we queried OSM features associated with keys such as \texttt{landuse}, \texttt{waterway}, \texttt{highway}, \texttt{building}, and \texttt{industrial}. The raw tags were saved and associated with their corresponding images. 

However, utilizing these raw OSM tags presented two primary challenges. First, the open-source nature of OSM leads to terminological inconsistencies (e.g., \texttt{Dam}, \texttt{dams}, \texttt{Private\_dam}, \texttt{weir}). Second, many tags denote fine-grained objects, such as a \texttt{clinic} or \texttt{gas station}, that are not resolvable at the 30-meter GSD of Landsat imagery. To mitigate these issues of inconsistency and scale mismatch, we designed a multi-tiered classification schema to map the raw tags into a controlled vocabulary. This schema normalizes the diverse tags into $25$ distinct land use categories appropriate for remote sensing analysis. The example mapping is detailed in Listing~\ref{osm-tag-mapping}.

\begin{listing}[ht]%
\caption{OpenStreetMap Tag Mapping Schema.}%
\label{osm-tag-mapping}%
\begin{lstlisting}[basicstyle=\ttfamily\small,breaklines=true,language=Python]
mapping_categories = {
"river_stream": r"\b(river|stream|creek|drain|canal|oxbow|tidal_channel)\b",
"wetland": r"\b(wetland|swamp|marsh|bog)\b",
"cropland": r"\b(farmland|farm|farmyard|paddock|cropland|orchard|"
r"plantation|vineyard|paddock|cotton_gin)\b",
"natural_vegetation": r"\b(forest|woodland|trees?|grassland|meadow|scrub|"
r"shrub)\b",
"urban_fabric": r"\b(residential|village|town|suburb|city|commercial|"
r"industrial|factory|warehouse|retail|parking|building|"
r"(ice_)?factory|bakery|university|hospital|school|"
r"clinic|shopping|mall|casino|pub|restaurant|hotel|"
r"clubhouse|office|gym|stadium|arena|terminal|depot)\b",
"road_corridor": r"\b(road|street|motorway|highway|primary|secondary|"
r"tertiary|trunk|track|service|path|railway|cycleway)\b",
...
}
\end{lstlisting}
\end{listing}

\subsubsection{Land cover reference.}

To supplement our imagery with land cover information, we leveraged the DEA Land Cover product. This dataset provides annually updated, pixel-level classifications for the Australian continent, derived from a full year of satellite observations. Within this product, each pixel is assigned to one of seven categories: \texttt{Cultivated Terrestrial Vegetation}, \texttt{Natural Terrestrial Vegetation}, \texttt{Natural Aquatic Vegetation}, \texttt{Artificial Surfaces}, \texttt{Natural Bare Surfaces}, \texttt{Water}, or \texttt{No Data}.

Crucially, this is a summary product; a pixel's classification for a given year reflects its predominant state over that entire period. For example, a pixel is labeled as \texttt{Water} only if it was observed as water for a significant majority of the year's clear observations. To integrate this information, we used our established AOI grid to extract a corresponding land cover map for each of our sampled Landsat images, ensuring precise spatial alignment between the visual data and its classification.

\subsection{Stage 2: Fine-tuning VLMs for Landsat Tasks}
\subsubsection{Region classification.}
\label{image-region-cls-prompt-and-review}

To generate a structured, region-based land cover classification for each image, we prompted the GPT-4o model. Crucially, rather than cropping the image into smaller segments, we provided the entire $256 \times 256$\,px true-color RGB patch as input. This approach was chosen to leverage the model's holistic geospatial understanding, allowing it to interpret features within the context of the entire scene. The detailed prompt guiding this process is provided in Listing~\ref{region-cls-prompt}.

The model was instructed to classify land cover for six distinct spatial regions: the top-left, bottom-left, top-right, bottom-right, and center, as well as for the overall scene. For each region, the land cover types were to be ordered by their coverage. The rationale behind this structured classification is to support a downstream captioning task. By providing the captioning model with this pre-analyzed geospatial information, we ensure it is aware of the dominant land cover in specific areas. This prevents the model from overlooking salient patterns and helps generate more accurate and contextually rich descriptions.

\begin{listing}[ht]%
\caption{Region Classification Prompt.}%
\label{region-cls-prompt}%
\begin{lstlisting}[basicstyle=\ttfamily\small,breaklines=true,language=Python]
prompt = (
"You are an advanced assistant for analyzing an optical satellite image. "
"Your role is using the information from image to accurate answers to the questions to the scene."
"Analyze an optical satellite image to classify land cover types. "
"Focus on six classifications: Cultivated Terrestrial Vegetation, Natural Terrestrial Vegetation, Natural Aquatic Vegetation, Artificial Surface, Natural Bare Surfaces, and Water. "
"Pay particular attention to Cultivated Terrestrial Vegetation, Artificial Surface, and Water.\n"

"Answer these questions:\n"
"1. What land cover classifications can be found in the image?\n"
"2. Divide the image into five sections: top-left, bottom-left, top-right, bottom-right, and center. "

"For each, list classifications in order of area occupied.\n")
\end{lstlisting}
\end{listing}

To assess the quality of region classification result, we initially turned to the DEA Land Cover product as a reference benchmark. However, using this product for direct, automatic validation presents a significant temporal mismatch. This Land Cover product is a yearly composite; for instance, its \texttt{Water} class designates areas that were inundated for a majority of the year, which may not align with the land cover captured in a single-date Landsat image. Similarly, distinguishing between classes like \texttt{Cultivated Terrestrial Vegetation} and \texttt{Natural Terrestrial Vegetation} often relies on seasonal patterns derived from time-series analysis, a context unavailable to a model interpreting a single image.

Given these inherent challenges with reference dataset, we opted for a manual verification process to create a reliable fine-tuning dataset. We sampled $2,722$ images and had human reviewers meticulously verify the accuracy and coverage order of the generated land cover classifications for each one. This rigorous process resulted in a high-quality, human-verified fine-tuning set of $2,722$ samples. We then partitioned this dataset into training (80\%), and test (20\%) sets for model development and evaluation. Using the resulting training set, we fine-tuned the GPT-4o model (as \textbf{region classification model}) via the OpenAI Playground. The training was conducted for 3 epochs with a batch size of 4, a learning rate multiplier of 2, and a random seed of 42, using the same inference prompt detailed in Listing~\ref{region-cls-prompt}.

To provide a comprehensive evaluation of our multi-label region-classification model, we employed a diverse suite of metrics designed to assess different facets of its performance, from exact-match accuracy to ranking quality.

\paragraph{Set-based Accuracy} For overall correctness, we use \textbf{Subset Accuracy}, the most stringent metric which considers a prediction correct only if the set of predicted labels is an exact match for the true labels. As a more forgiving alternative, we also report the \textbf{Jaccard Index} (intersection over union), which measures the similarity between the predicted and true label sets, penalizing both missing and incorrect labels.

\paragraph{Label-based Accuracy} To evaluate performance on a per-label basis, we utilize the standard metrics of \textbf{Precision} (the fraction of predicted labels that are correct, measuring exactness), \textbf{Recall} (the fraction of true labels that were successfully predicted, measuring completeness), and the \textbf{F1-score}, which provides a balanced measure as the harmonic mean of Precision and Recall.

\paragraph{Ranking Quality} Since our task requires the model to order labels by coverage, we assess ranking quality using two key metrics. \textbf{Label-Ranking Average Precision (LRAP)} averages over each ground-truth label the proportion of higher-ranked labels that are also true. \textbf{Normalized Discounted Cumulative Gain (nDCG)} evaluates the ranked list by assigning higher scores to correct labels placed earlier in the prediction, providing a granular measure of ranking effectiveness.

\paragraph{Loss-based Metrics} Finally, we report two loss-based metrics, presented as $(1 - loss)$ so that higher values consistently indicate better performance. \textbf{1 - Hamming Loss} reflects the fraction of correctly predicted labels out of the total number of labels. \textbf{1 - Ranking Loss} measures the fraction of relevant-irrelevant label pairs that are correctly ordered by the model.

\subsubsection{Image captioning.}
\label{image-caption-prompt}

For our image captioning task, we constructed another dedicated fine-tuning dataset using a methodology analogous to our region classification approach. We first employed a prompted GPT-4.1 model to generate initial draft captions. These captions then underwent a rigorous manual verification process, where human reviewers confirmed their accuracy and relevance, resulting in a high-quality, human-verified dataset for fine-tuning.

During the caption generation inference, we provided the VLM with a rich set of VLM inputs: the Landsat image itself, the structured land cover classifications from our fine-tuned GPT-4o model, and the standardized land use information derived from OSM tags (as per the schema in Listing~\ref{osm-tag-mapping}). A critical component of our prompt engineering was to instruct the model to cross-validate these data sources to prevent hallucinations. For instance, if the land cover data indicated no water in the scene, the VLM was directed to ignore any water-related land use tags from OSM. This strategy mitigates errors arising from potentially outdated or mismatched OSM tags. The detailed prompt for this task is provided in Listing~\ref{img-cap-prompt}.

\begin{listing}[ht]%
\caption{Image Captioning Prompt.}%
\label{img-cap-prompt}%
\begin{lstlisting}[basicstyle=\ttfamily\small,breaklines=true,language=Python]
system_prompt = """Generate a detailed and concise caption from an optical satellite image using provided metadata.

1. Please use image content and land cover information to cross-validate land use information. Please use verifed land use information to finish the caption.
2. Identify all visible water bodies; mention locations and relative sizes.
3. Distinguish dominant land cover in each area, specify approximate extent or pattern if possible.
4. Identify and size artificial areas ("small town", "city") and reference exact locations when relevant.
5. Describe visible urban features only if seen or confirmed in metadata.
6. Describe road corridors, specifying directions and links to urban areas, if visible.
7. Include any spatial references for features (top, bottom, left, right, center).
8. Summarize the balance and dominance between bare and vegetated surfaces.
9. Use the land use metadata to offer insights on overall landscape use."""

user_prompt = (
"The following are the metadata to this satellite image:\n "
"Land Cover Information:\n "
f"{lc_formatted}\n"
"Land Use Information:\n "
f"{landuse}"
)

\end{lstlisting}
\end{listing}

To establish a gold-standard benchmark and ensure the quality of our captions, we implemented a rigorous manual verification process. The resulting set of human-verified image-caption pairs serves a dual role: it provides the primary dataset for fine-tuning our image captioning model and simultaneously functions as the official test set for our \textsc{Landsat30-AU-Cap} benchmark.

To facilitate this detailed review, we first enhanced the visual clarity of the imagery by pansharpening the 30-meter GSD source images to 15-meter GSD using Landsat's panchromatic band. Our human reviewers then evaluated each caption on a sentence-by-sentence basis. For each sentence, the task was a binary decision: to \texttt{keep} it if its claim was visually verifiable in the enhanced image, or to \texttt{delete} it otherwise. In cases where the image alone was insufficient for confirmation, reviewers were authorized to consult high-resolution, third-party sources like Google Earth to make a final determination. This process generated not only the clean caption dataset but also a log of \textit{image-sentence-decision} triplets.

\begin{figure*}[t]
  \centering
  \begin{subfigure}[t]{0.48\textwidth}
    \centering
    \includegraphics[width=\linewidth]{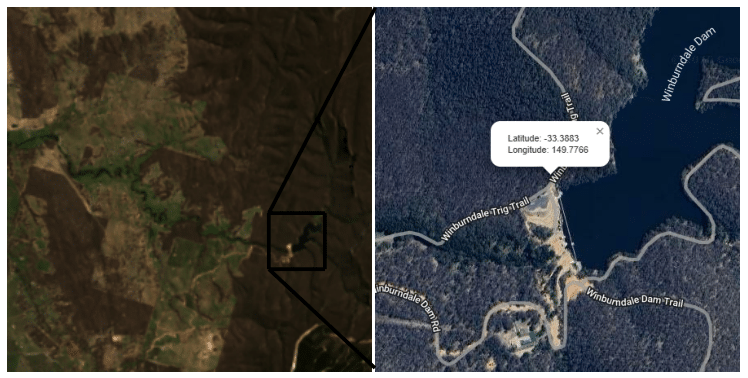}
    \caption{A small dam in the lower right. Decision: Keep.}
    \label{fig:cap-vqa-human-verify-dam}
  \end{subfigure}\hfill
  \begin{subfigure}[t]{0.48\textwidth}
    \centering
    \includegraphics[width=\linewidth]{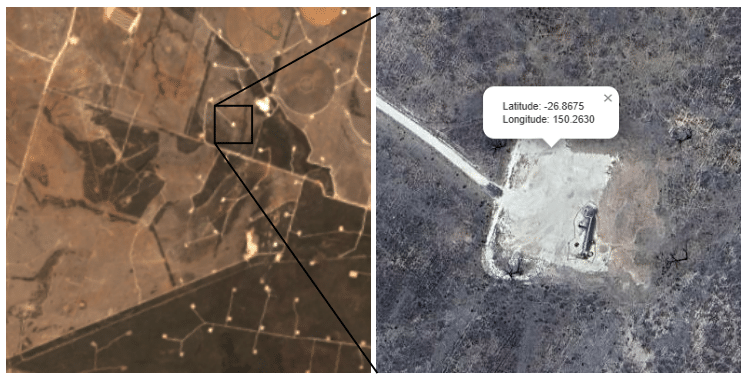}
    \caption{Grid of white oil or gas well pads. Decision: Keep.}
    \label{fig:cap-vqa-human-verify-gas-well}
  \end{subfigure}

  \begin{subfigure}[t]{0.48\textwidth}
    \centering
    \includegraphics[width=\linewidth]{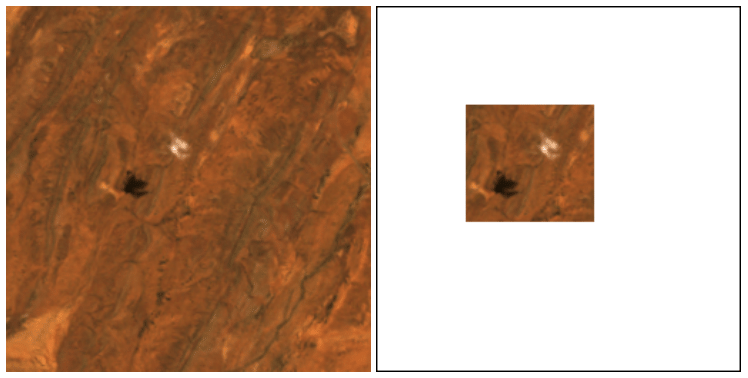}
    \caption{Spatial relationship between water body and bare surface? Fix Answer: no water body.}
    \label{fig:cap-vqa-human-verify-cloud}
  \end{subfigure}\hfill
  \begin{subfigure}[t]{0.48\textwidth}
    \centering
    \includegraphics[width=\linewidth]{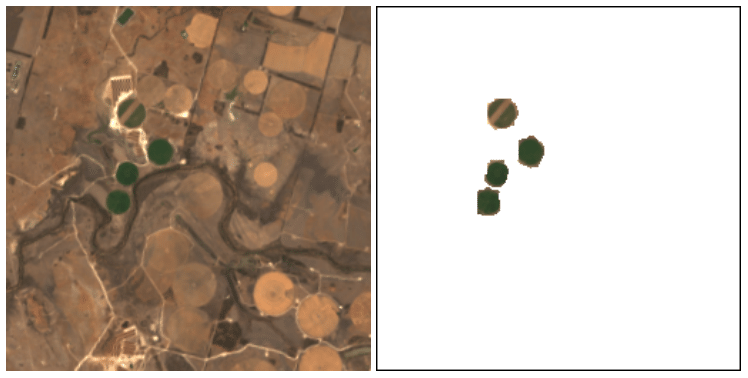}
    \caption{How many green circular irrigated fields? Fix: from two to four.}
    \label{fig:cap-vqa-human-verify-number}
  \end{subfigure}

    \begin{subfigure}[t]{0.48\textwidth}
    \centering
    \includegraphics[width=\linewidth]{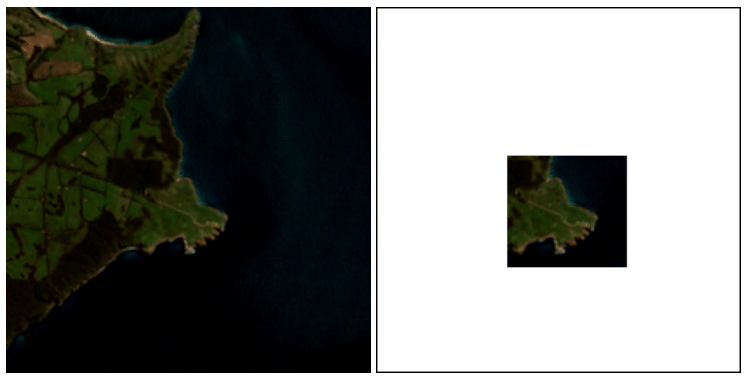}
    \caption{How many docks in the image? Fix: from three to zero.}
    \label{fig:cap-vqa-human-verify-dock}
  \end{subfigure}\hfill
  \begin{subfigure}[t]{0.48\textwidth}
    \centering
    \includegraphics[width=\linewidth]{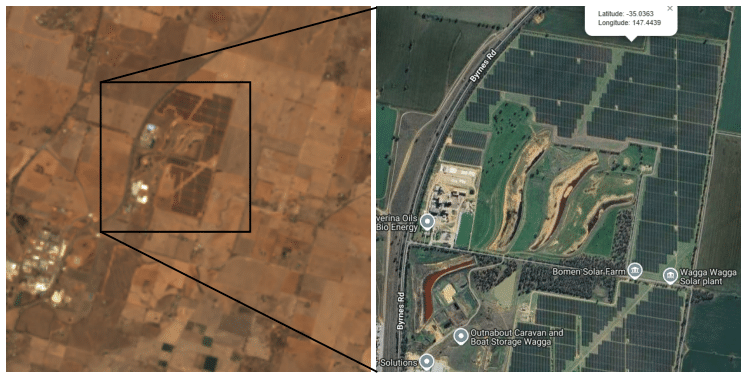}
    \caption{Verified Q: Which object can be found from the image? Verified A: solar farm.}
    \label{fig:cap-vqa-human-verify-solar-farm}
  \end{subfigure}

  \caption{More caption and VQA human-verification examples.}
  \label{fig:cap-vqa-human-verify-more}
\end{figure*}

This level of review was particularly critical for ambiguous or difficult-to-discern features. For instance, as shown in Fig.~\ref{fig:cap-vqa-human-verify-dam}, a caption referenced a dam that was not obvious in the base imagery; external verification with Google Maps confirmed its presence. Similarly, in the case presented in Fig.~\ref{fig:cap-vqa-human-verify-gas-well}, a caption identified a grid-like pattern as gas well pads. This claim, difficult to verify visually, was confirmed by cross-referencing high-resolution imagery and local information.

This human verification process yielded a curated dataset of $1,005$ high-quality image-caption pairs, which we partitioned into training (70\%), validation (15\%), and test (15\%) sets. We then fine-tuned the GPT-4.1 model (as \textbf{image captioning model}) on the OpenAI Playground using this training data. The fine-tuning was conducted for 3 epochs with a batch size of 1, a learning rate multiplier of 2, and a random seed of 42 for reproducibility. This process utilized the same inference prompt as detailed in Listing~\ref{img-cap-prompt}.

\subsubsection{Caption review.}

To automate caption validation at scale and to mitigate potential biases from relying on a single model family, we trained a dedicated open-source VLM (Qwen2.5-VL-7B) to function as a \textbf{caption review model}. The model's task was to review each sentence of a generated caption against the corresponding image and determine whether it should be \texttt{kept} or \texttt{deleted} based on its visual accuracy. The detailed prompt guiding this review task is provided in Listing~\ref{review-sentence-prompt}.

The training data for this review model was a direct byproduct of our manual verification process. The human-verified decisions from the image-captioning dataset construction yielded a new dataset of 9,440 image-sentence-decision triplets, where each sentence was labeled with a human-confirmed \texttt{keep} or \texttt{delete} action. We partitioned this dataset using a 70/15/15 split for training, validation, and testing, respectively. We then performed a full-parameter fine-tuning of the Qwen model on this data. The model was trained for three epochs with a batch size of 24. We used an Adam optimizer with a cosine learning rate schedule, starting at $2 \times 10^{-5}$ with a warm-up phase over the first 6\% of steps. The configuration also included an L2 weight decay of $1 \times 10^{-6}$ and an Adam $\beta_2$ value of 0.999. The training utilized bfloat16 mixed-precision and was conducted on a server equipped with eight NVIDIA L4-24G GPUs.

\begin{listing}[ht]%
\caption{Caption Review.}%
\label{review-sentence-prompt}%
\begin{lstlisting}[basicstyle=\ttfamily\small,breaklines=true,language=Python]
system_prompt = """Check the given image and its corresponding caption (a single sentence) to determine if the caption accurately reflects the content of the image. Respond with either "delete" if the caption does not match the image content or "keep" if it does.

# Steps
1. Analyze the content of the provided image to understand its main elements and context.
2. Read the given caption and evaluate its accuracy and relevance to the image content.
3. Decide whether the caption accurately represents the image.
4. Respond accordingly with "delete" or "keep".

# Notes
- The caption should be a clear and direct reflection of the image's primary content.
- Consider the main focus of the image, including any prominent objects, actions, or emotions.
- "Keep" the caption if it correctly and completely represents the image without ambiguity. Otherwise, choose "delete"."""

user_prompt = f"The given caption: {given_caption}\n"
\end{lstlisting}
\end{listing}

\subsection{Stage 3: Multi-Stage Caption and VQA Generation}

\subsubsection{Caption refinement.}

To enrich the initial captions generated by our fine-tuned GPT-4.1 model, we designed and implemented a sequential, two-stage refinement pipeline. This pipeline leverages a prompted Qwen2.5-VL-7B model to systematically incorporate salient visual details that were initially omitted.

The pipeline operates as follows:
\begin{enumerate}
    \item \textbf{Missing Object:} In the first stage, the Qwen model is prompted to identify and add a description of a prominent object or pattern that is visible in the image but absent from the original caption.
    \item \textbf{Missing Connection:} In the second stage, the model analyzes the newly augmented caption from the previous step. It is prompted to articulate a spatial or contextual relationship between objects that was not previously described, further increasing the caption's descriptive depth.
\end{enumerate}

Crucially, the input for the \textbf{Missing Connection} stage is the output from the \textbf{Missing Object} stage, allowing for progressively more complex descriptions. We observed that this enrichment process can occasionally introduce redundant phrases. However, these redundancies are subsequently addressed and removed by our caption review model (a fine-tuned QWen2.5-VL-7B model), which serves as the final quality control step in our generation pipeline. The detailed prompts for each refinement stage are provided in Listing~\ref{add-missing-object-prompt} and Listing~\ref{add-missing-connection-prompt}.

\begin{listing}[ht]%
\caption{Add Missing Object Prompt.}%
\label{add-missing-object-prompt}%
\begin{lstlisting}[basicstyle=\ttfamily\small,breaklines=true,language=Python]
system_prompt = """You are an advanced assistant specializing in analyzing optical satellite images.  You will get a caption about one image. Your task is to find and describe special missing patterns or objects that appear in the image but not in caption. 

Only describe what is clearly visible - do NOT mention anything that is absent or not shown in the image. Avoid making statements about what is not present. 

Mention only one key missing object or pattern that is clearly visible in the image but not in the caption; keep it concise and ideally contained within a single sentence.

This will instruct me to avoid referencing absent features in my responses."""

user_prompt = f"The given caption: {caption}\n"

\end{lstlisting}
\end{listing}

\begin{listing}[ht]%
\caption{Add Missing Connection Prompt.}%
\label{add-missing-connection-prompt}%
\begin{lstlisting}[basicstyle=\ttfamily\small,breaklines=true,language=Python]
system_prompt = """You are an advanced assistant specializing in analyzing optical satellite images.  You will get a caption about one image. Your task is to find and describe special missing connections between objects that appear in the image but not in caption. 

Only describe what is clearly visible - do NOT mention anything that is absent or not shown in the image. Avoid making statements about what is not present. 

Mention only one key missing connection or relationship that is clearly visible in the image but not in the caption; keep it concise and ideally contained within a single sentence.

This will instruct me to avoid referencing absent features in my responses."""

user_prompt = f"The given caption: {caption}\n"

\end{lstlisting}
\end{listing}

To comprehensively evaluate the quality of the generated captions, we employ a suite of metrics that assess fluency, semantic fidelity, and factual accuracy against our human-verified reference captions.

\paragraph{Fluency and N-gram Matching}
We use \textbf{BLEU-4} to measure n-gram precision. This metric calculates the overlap of 4-gram (four-word) sequences between the generated caption and the reference captions, providing a measure of grammatical correctness and fluency. While a standard metric, it does not capture semantic similarity beyond exact word matches.

\paragraph{Semantic Fidelity}
To evaluate how well the caption captures the meaning of the scene, we use two advanced metrics. \textbf{SPIDEr} is a composite metric that combines the strengths of SPICE, which evaluates the alignment of objects, attributes, and relations using a scene graph, and CIDEr, which measures n-gram similarity while weighting less common but more informative phrases. We also use \textbf{BERTScore-F1}, which leverages contextual embeddings from BERT to measure the semantic similarity between tokens. Unlike n-gram-based metrics, BERTScore can recognize synonyms and paraphrasing (e.g., \texttt{river} and \texttt{waterway}), providing a more robust measure of a caption's fidelity to the reference text's meaning.

\paragraph{Hallucination Measurement}
A critical aspect of our evaluation is the quantification of object hallucination, for which we use the \textbf{CHAIR} (Caption-Hallucination-Assessment-with-Image-Representations) metric. We report two versions: \textbf{CHAIR-s} (sentence-level), which calculates the percentage of sentences containing at least one hallucinated object, and \textbf{CHAIR-i} (instance-level), which calculates the percentage of all object instances mentioned in a caption that are hallucinated. 

To implement this, our methodology involves a two-step process of extraction and normalization. First, we extract all potential object nouns from each generated caption using a prompted GPT-4.1 model, with the specific prompt detailed in Listing~\ref{extract-object-prompt}.

However, raw nouns can be inconsistent due to synonyms (e.g., \texttt{causeway} versus \texttt{bridge}). To enable a fair and standardized comparison, we then normalize these terms by mapping them into one of 43 predefined object categories using a custom schema; an example of this mapping is provided in Listing~\ref{extract-objects-mapping}. These standardized object categories are then cross-referenced against the ground-truth list of objects present in the image. Finally, to align these metrics with our other evaluations, we report the hallucination scores as \textbf{1-CHAIR-s} and \textbf{1-CHAIR-i}. In this format, a higher score indicates a lower hallucination rate and thus represents better performance.

\begin{listing}[ht]%
\caption{Extract Objects from Caption Prompt.}%
\label{extract-object-prompt}%
\begin{lstlisting}[basicstyle=\ttfamily\small,breaklines=true,language=Python]
system_prompt = """Extract the key objects directly from the provided caption. These objects must explicitly appear in the caption.

# Steps
1. Carefully read the provided caption, identifying each object explicitly mentioned.
2. Cross-check each identified object to ensure it directly appears in the caption and falls under categories related to earth observation such as natural features, human-made structures, or land use areas.

# Output Format
- Return a JSON array containing strings of all identified key objects relevant to earth observation, directly extracted from the caption.

# Examples
**Input:** "The image shows a large river bending through a dense forest with a small urban area visible on the horizon."
**Output:** ["river", "forest", "urban area"]

user_prompt = f"The given caption: {given_caption}\n"
\end{lstlisting}
\end{listing}

\begin{listing}[ht]%
\caption{Extracted Object Mapping Schema Example.}%
\label{extract-objects-mapping}%
\begin{lstlisting}[basicstyle=\ttfamily\small,breaklines=true,language=Python]
OrderedDict(
[
    ("Bridge",        r"\b(bridg[-\w]*|causeway[-\w]*)\b"),
    ("Dam",           r"\b(dam[-\w]*|weirs?)\b"),
    ("Harbor",        r"\b(dock[-\w]*|harbor|harbour?[-\w]*|port[-\w]*|marina[-\w]*|jetty[-\w]*|pier[-\w]*)\b"),
    ("Airport",       r"\b(airport[-\w]*|air\s?strip[-\w]*|airfield[-\w]*|runway[-\w]*)\b"),
    ("Golf Course",   r"\b(golf\s?course[-\w]*)\b"),
    ("Solar Farm",    r"\b(solar[-\w]*)\b"),
    ("Lagoon",        r"\b(lagoon[-\w]*)\b"),
    ("Volcanic Crater", r"\b(volcanic\s?crater[-\w]*)\b"),
    ("Green House",   r"\b(green\s?house[-\w]*|greenhouse[-\w]*)\b"),
    ("Delta",         r"\b(delta[-\w]*)\b"),
    ...
]
)
\end{lstlisting}
\end{listing}

\subsubsection{VQA generation.}
\label{generate-vqa-from-caption}


To systematically probe different facets of VLM performance, we designed eight targeted VQA categories. These are grouped into two sets: advanced tasks designed to test for robustness against common remote sensing challenges, and foundational visual comprehension tasks derived directly from our verified captions.

\paragraph{Advanced Remote Sensing and Robustness Tasks}
Building on this foundation, we introduced four categories targeting complex, domain-specific challenges inherent to remote sensing analysis. The \textbf{Agro-Phenology Reasoning (APR)} task evaluates a model's ability to infer cropping seasons from a single image, with ground-truth labels derived from capture date and location. The \textbf{Cloud-Occlusion Assessment (COA)} task tests the ability to distinguish clouds from visually similar features like salt lakes or building reflections. The \textbf{Fine-Object Detectability (FOD)} task directly probes a model's understanding of scale and its propensity for hallucination by asking about objects, such as fences or powerlines, that are too small to be resolved at 30-meter GSD. Finally, the \textbf{Urban-Scale Recognition (USR)} task assesses whether a model can accurately classify settlements as a large city, a small town, or a rural area.

\paragraph{Foundational Visual Comprehension Tasks}
Leveraging our detailed, human-verified captions, we prompted GPT-4.1 to generate questions for four core categories that assess a model's fundamental understanding of image content. These include: \textbf{Macro-Object Presence (MOP)}, which tests for the existence of major objects; \textbf{Numerosity (NUM)}, which evaluates counting abilities; \textbf{Spatial-Relation Inference (SRI)}, which probes the understanding of positional relationships between objects; and \textbf{Dominant Land-Cover Classification (DLC)}, which assesses scene-level classification.

Recognizing that these model-generated pairs could contain factual errors due to hallucination, we subjected the entire set to a rigorous human verification process. This manual review served two primary functions: first, to correct factual inaccuracies in the generated content, and second, to strategically increase the dataset's difficulty by introducing more challenging and nuanced questions. We also mask all geolocation and capture time information from image URLs. This is a critical step to prevent sensitive metadata leakage to VLMs.

\paragraph{Correcting Factual Inaccuracies}
Reviewers first rectified factual errors made by the model. For instance, in the case shown in Fig.~\ref{fig:cap-vqa-human-verify-cloud}, the model had misidentified a cloud shadow as a water body; the reviewer corrected the question's options and designated \texttt{no water body} as the answer. Similarly, another reviewer corrected an incorrect object count from \texttt{two} to \texttt{four} (Fig.~\ref{fig:cap-vqa-human-verify-number}), ensuring the dataset's numerical accuracy.

\paragraph{Enhancing Dataset Difficulty}
Second, reviewers actively enhanced the dataset's difficulty by introducing adversarial examples. This often involved creating questions that test for common visual misclassifications. In Fig.~\ref{fig:cap-vqa-human-verify-dock}, which depicts cliffs visually similar to docks, a reviewer crafted the question, \texttt{How many docks are in the image?}, with the correct answer being \texttt{zero}. Reviewers also added questions about salient objects the model had overlooked entirely. For example, when the model failed to generate any questions about a prominent solar farm (Fig.~\ref{fig:cap-vqa-human-verify-solar-farm}), a reviewer manually added a question about its presence to better benchmark VLM perception for such structures.

\begin{listing*}[ht]%
\caption{Generate VQAs Prompt.}%
\label{generate-vqa-prompt}%
\begin{lstlisting}[basicstyle=\ttfamily\small,breaklines=true,language=Python]
system_prompt = """You are an AI that generates multiple-choice questions based on a given image and a key object list. Your questions should focus on four aspects: scene/land-cover identification, object presence, counting, and spatial relations.

Instructions:

Given an input of an image and its associated key object list, follow these steps:
Scene/Land-Cover Identification:
Analyze the environment or landscape in the image.
Generate a multiple-choice question that helps identify or classify the scene type.

Object Presence:
Assess which key objects from the list are visible in the image.
Create a question to confirm or deny the presence of these objects.

Counting:
Count the number of specified key objects visible in the image.
Formulate a question to verify this count.

Spatial Relation:
Evaluate the spatial relationships between notable objects.
Generate a question to describe or identify these relationships.

Output Format:
Produce your output as a Python dictionary with the following structure:
{
    "questions": [
        {
            "type": "scene_land_cover",
            "question": "<question_text>",
            "choices": ["<choice_1>", "<choice_2>", "<choice_3>", "<choice_4>"],
            "answer": "<correct_choice>"
        },
        {
            "type": "object_presence",
            "question": "<question_text>",
            "choices": ["<choice_1>", "<choice_2>", "<choice_3>", "<choice_4>"],
            "answer": "<correct_choice>"
        },
        {
            "type": "counting",
            "question": "<question_text>",
            "choices": ["<choice_1>", "<choice_2>", "<choice_3>", "<choice_4>"],
            "answer": "<correct_choice>"
        },
        {
            "type": "spatial_relation",
            "question": "<question_text>",
            "choices": ["<choice_1>", "<choice_2>", "<choice_3>", "<choice_4>"],
            "answer": "<correct_choice>"
        }
    ]
}
Each question dictionary contains:
    "type": The aspect being questioned.
    "question": The question text.
    "choices": A list of four answer choices (including three distractors and one correct answer in random order).
    "answer": The correct answer (as it appears in "choices")."""

user_prompt = f"object list: [{object_list}]"

\end{lstlisting}
\end{listing*}

\section{Landsat30-AU Dataset}

\subsection{Landsat30-AU-Cap.}

\begin{figure*}[t]
  \centering

\begin{subfigure}[t]{0.65\textwidth}
    \centering
    \includegraphics[width=\linewidth]{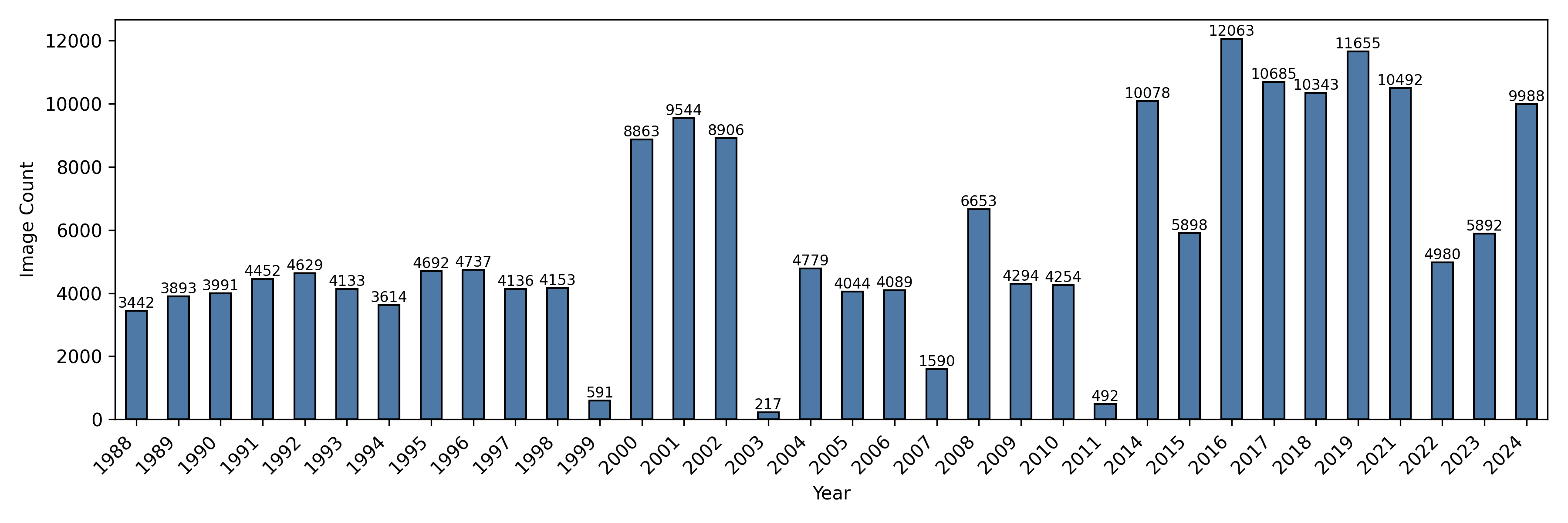}
    \caption{Temporal distribution of \textsc{Landsat30-AU-Cap}.}
    \label{fig:capture_year_distribution}
  \end{subfigure}
  \hfill
 \begin{subfigure}[t]{0.31\textwidth}
    \centering
    \includegraphics[width=\linewidth]{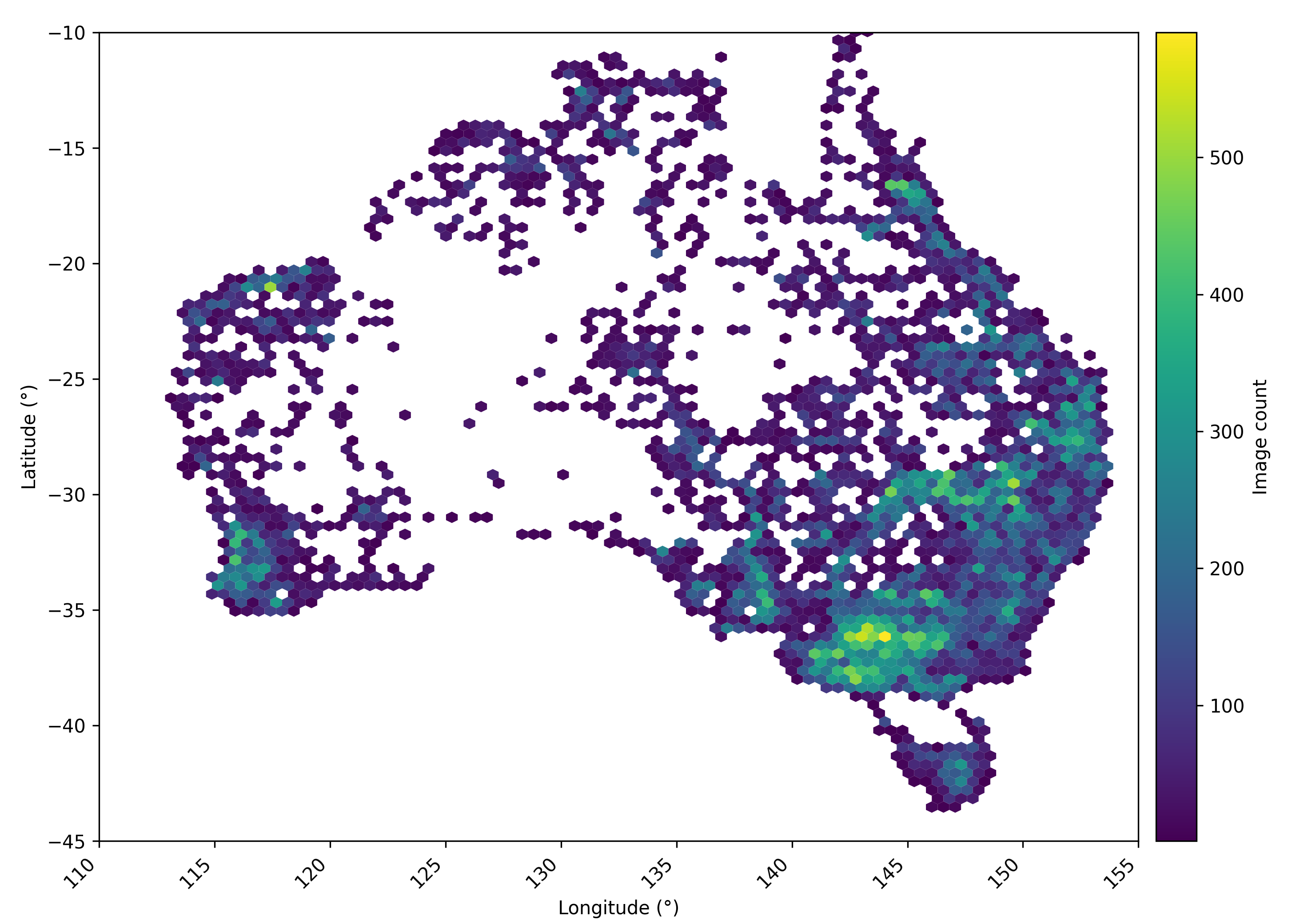}
    \caption{Spatial distribution of \textsc{Landsat30-AU-Cap}.}
    \label{fig:caption_geospatial_hexbin}
  \end{subfigure}
  \begin{subfigure}[t]{0.50\textwidth}
    \centering
    \includegraphics[width=\linewidth]{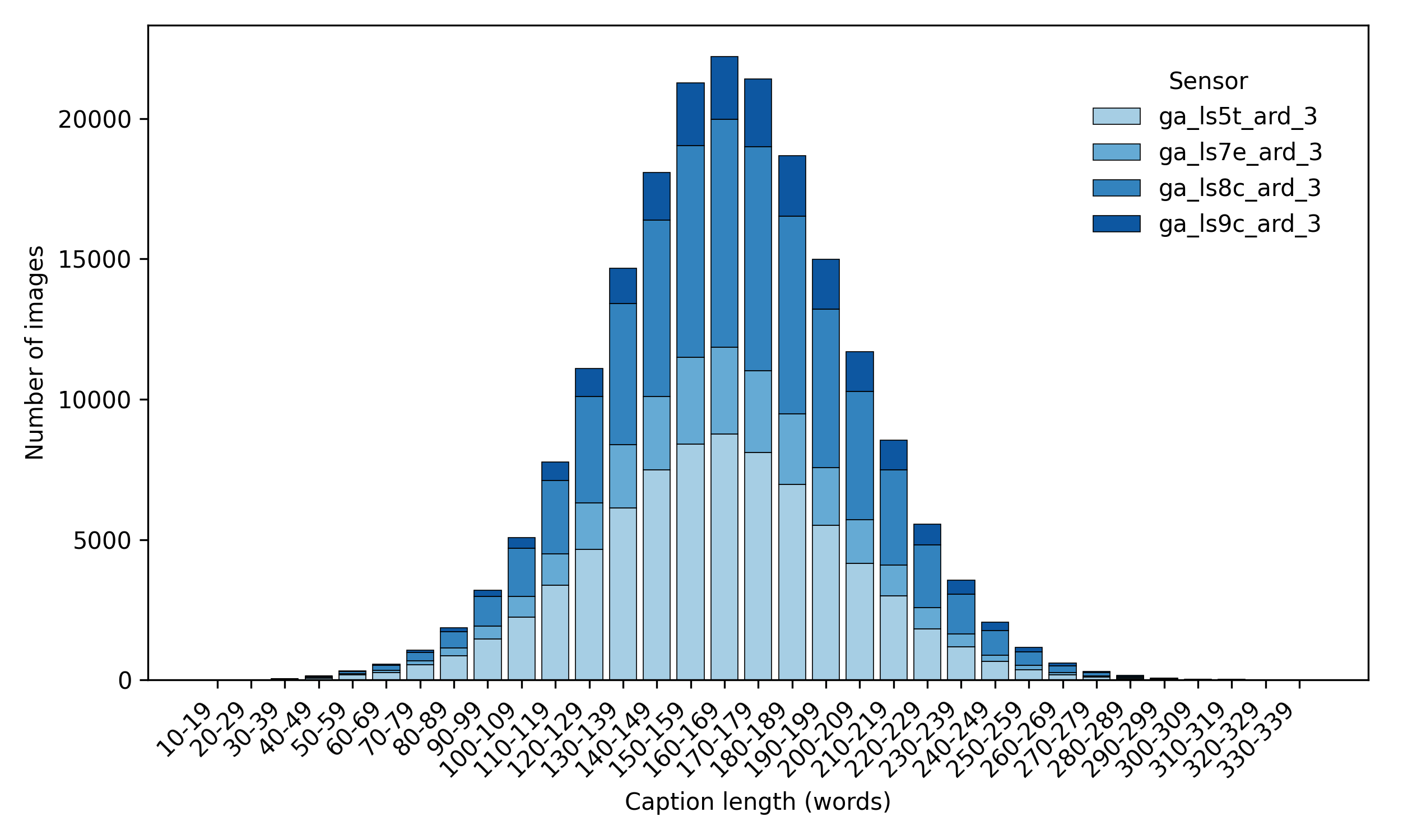}
    \caption{Caption-length distribution by Landsat satellites.}
    \label{fig:caption_length_by_sensor_stacked}
  \end{subfigure}
  \hfill
      \begin{subfigure}[t]{0.38\textwidth}
    \centering
    \includegraphics[width=\linewidth]{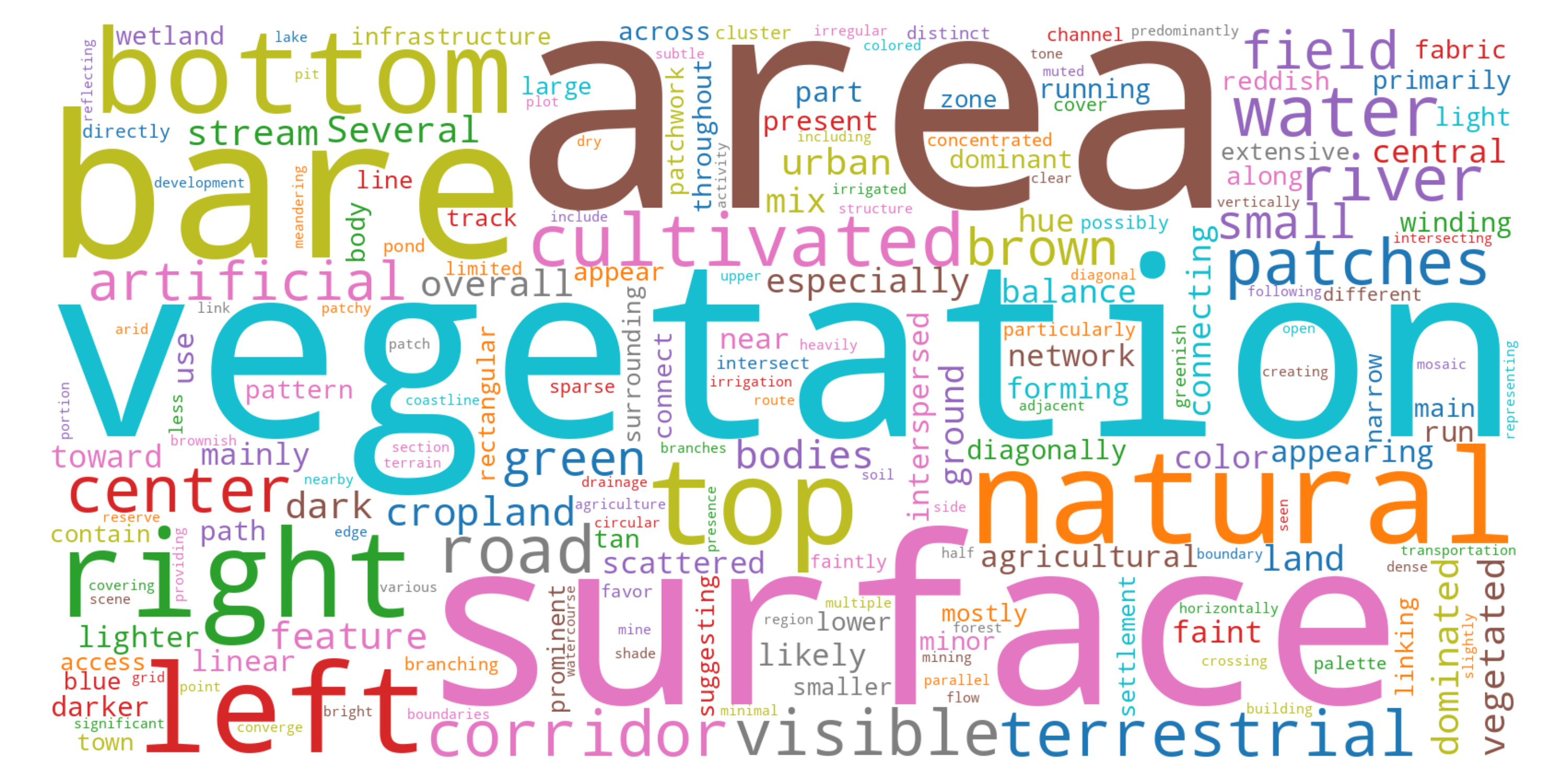}
    \caption{\textsc{Landsat30-AU-Cap} Word Cloud.}
    \label{fig:caption-wordcloud}
  \end{subfigure}

  \caption{Dataset statistics for \textsc{Landsat30-AU-Cap}.}
  \label{fig:detail-caption-profile}
\end{figure*}

Fig.~\ref{fig:detail-caption-profile} presents a comprehensive statistical profile of the Landsat30-AU dataset. The dataset's temporal distribution, illustrated in Fig.~\ref{fig:capture_year_distribution}, spans nearly four decades from 1988 to 2024. The geospatial distribution of the dataset, presented in Fig.~\ref{fig:caption_geospatial_hexbin}, reveals a clear clustering of data points in eastern and southwestern Australia, which aligns with the nation's most populated and agriculturally significant regions. Furthermore, Fig.~\ref{fig:caption_length_by_sensor_stacked} shows that the caption lengths approximate a normal distribution, demonstrating considerable descriptive depth. Lastly, the dominant terms in the word cloud (Fig.~\ref{fig:caption-wordcloud}), such as \texttt{vegetation}, \texttt{bare surface}, and \texttt{water}, highlight the dataset's focus on land cover and environmental features appropriate for Landsat's 30-meter GSD.

\subsection{Landsat30-AU-VQA.}

\begin{figure*}[t]
  \centering
    \begin{subfigure}{0.6\textwidth}
    \centering
    \includegraphics[width=\linewidth]{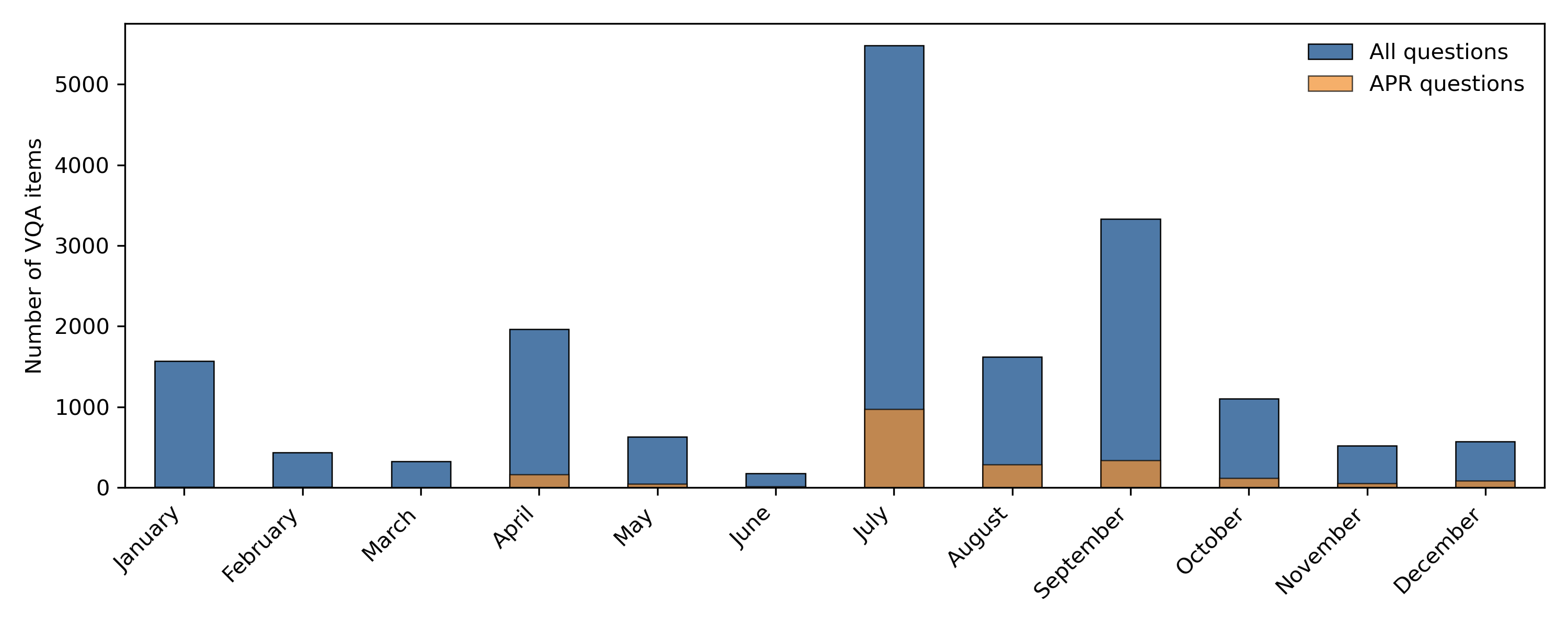}
    \caption{Temporal distribution of \textsc{Landsat30-AU-VQA}.}
    \label{fig:vqa-month-dist}
  \end{subfigure}
  \hfill
  \begin{subfigure}[t]{0.38\textwidth}
    \centering
    \includegraphics[width=\linewidth]{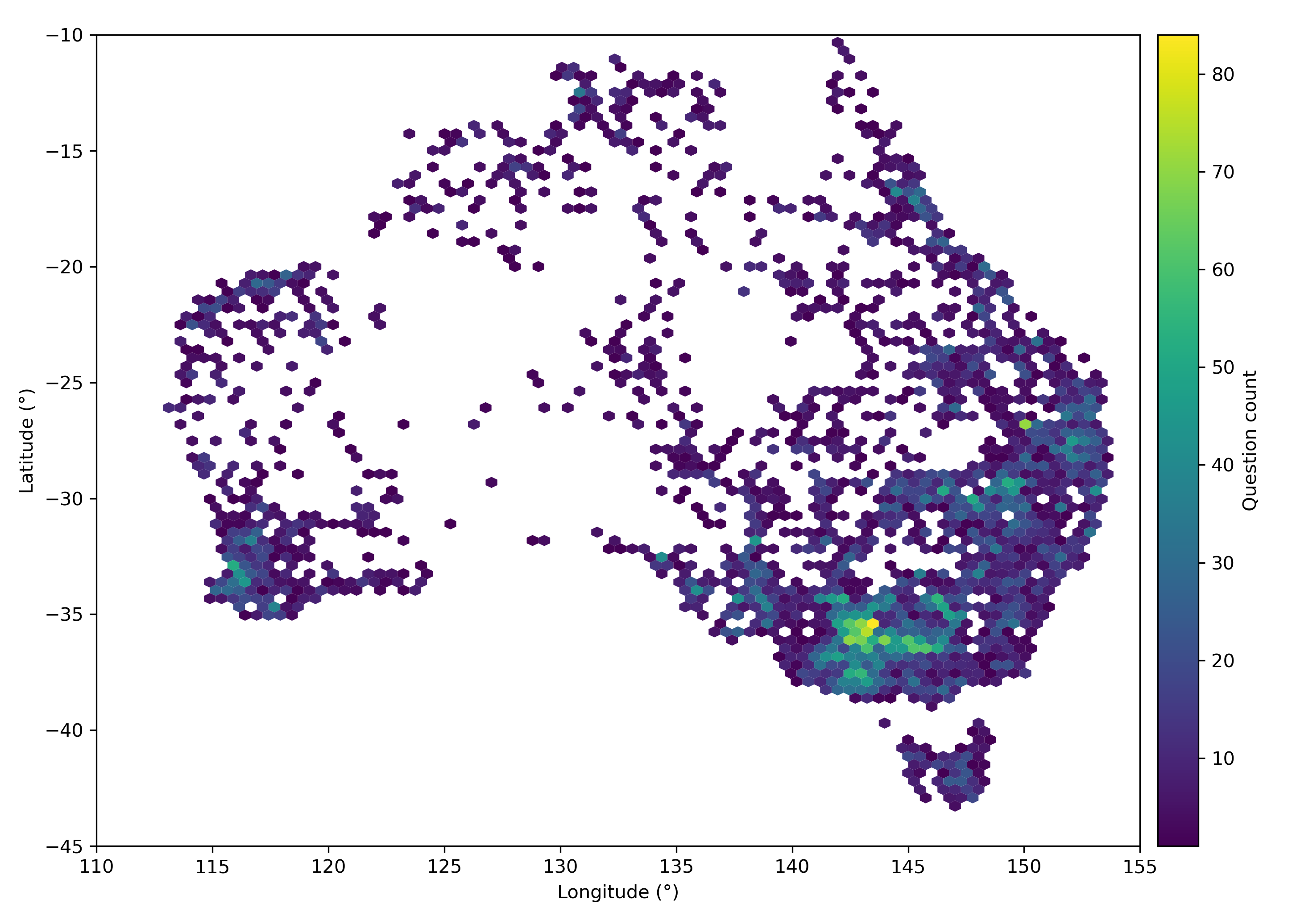} 
    \caption{Spatial distribution of \textsc{Landsat30-AU-VQA}.}
    \label{fig:vqa-spatial-dist}
  \end{subfigure}

  \begin{subfigure}[t]{0.23\textwidth}
    \centering
    \includegraphics[width=\linewidth]{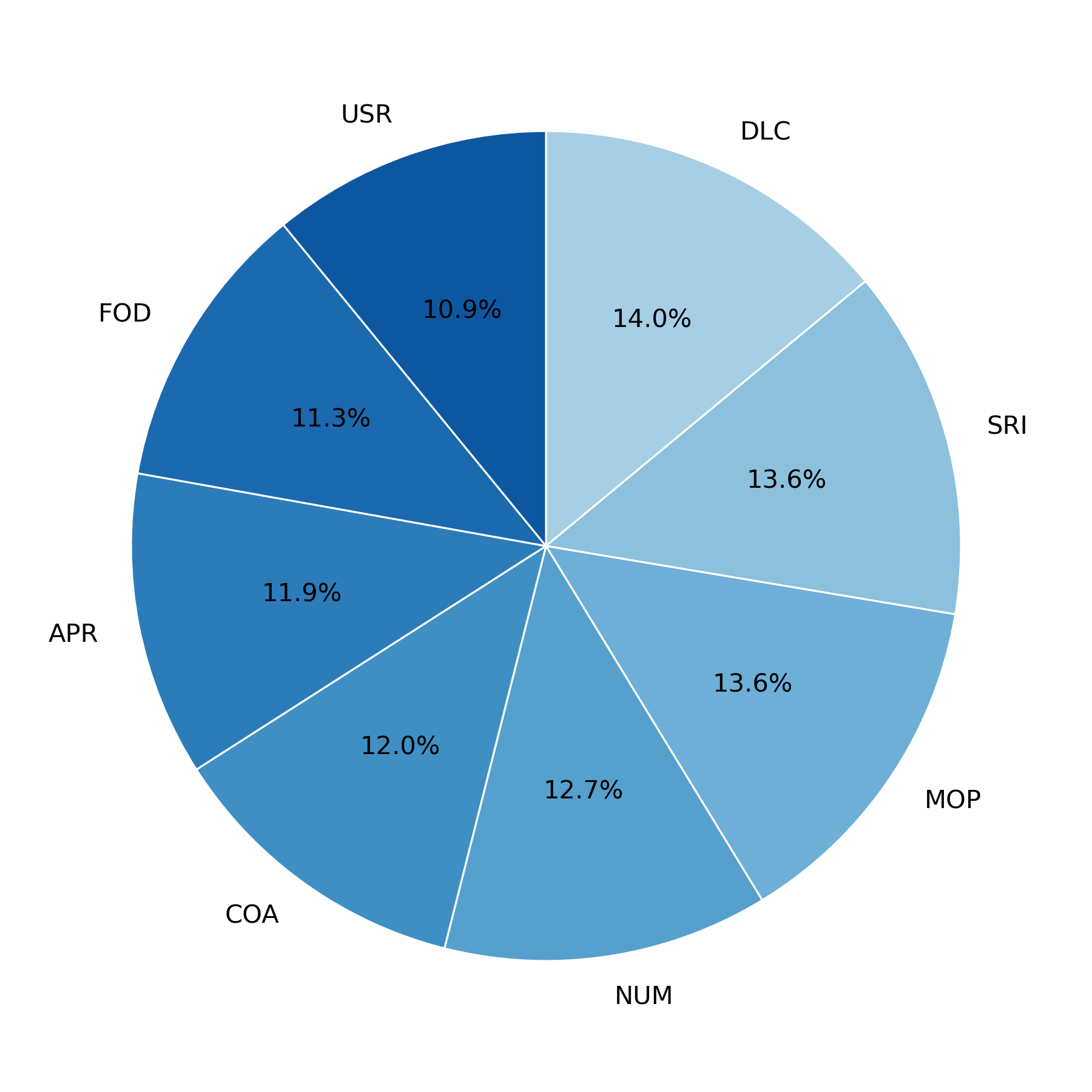} 
    \caption{VQA categories distribution.}
    \label{fig:vqa-qa-chart}
  \end{subfigure}
  \hfill
    \begin{subfigure}[t]{0.25\textwidth}
    \centering
    \includegraphics[width=\linewidth]{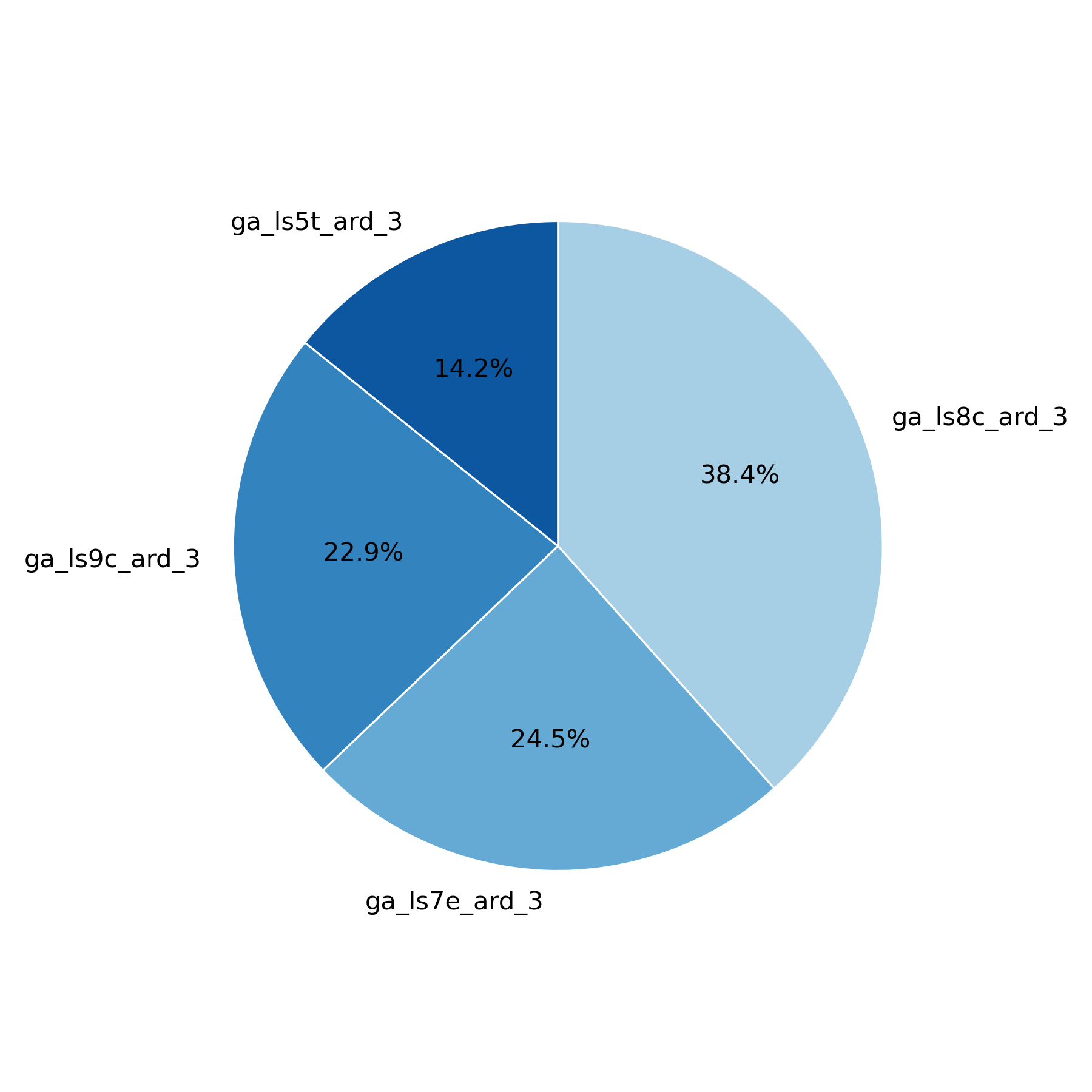} 
    \caption{VQA sensor distribution.}
    \label{fig:vqa-sensor-chart}
  \end{subfigure}
  \hfill
    \begin{subfigure}[t]{0.38\textwidth}
    \centering
    \includegraphics[width=\linewidth]{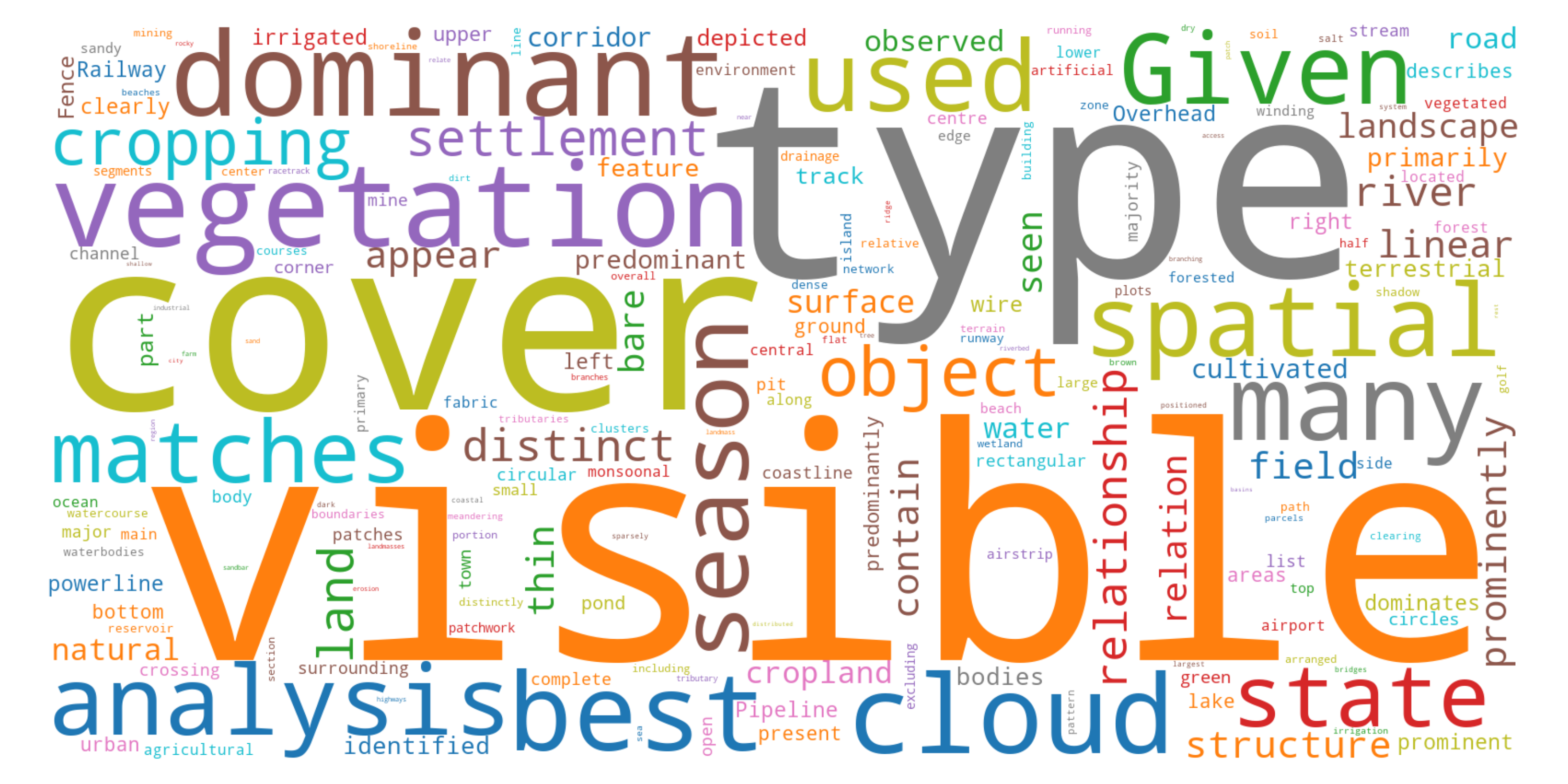} 
    \caption{\textsc{Landsat30-AU-VQA} Questions Word Cloud.}
    \label{fig:vqa-wordcloud}
  \end{subfigure}

  \caption{Dataset statistics for \textsc{Landsat30-AU-VQA}.}
  \label{fig:vqa_dataset_profile}
\end{figure*}

Fig.~\ref{fig:vqa_dataset_profile} provides a comprehensive statistical profile of the \textsc{Landsat30-AU-VQA} dataset, highlighting its key attributes across temporal, thematic, categorical, and spatial dimensions. 

The temporal profile of the dataset, detailed in Fig.~\ref{fig:vqa-month-dist}, highlights the successful inclusion of questions across all four seasons, which was a critical goal for the \textbf{Agro-Phenology Reasoning (APR)} task. We deliberately weighted sampling toward April-September, coinciding with the sowing-to-grain-fill window for Australia's main winter cereals. Harvest imagery (Nov-Dec) and summer-crop stages (Dec-May) are comparatively under-represented because of persistent wet-season cloud in the tropical north and data-acquisition limits. This skew mirrors real-world observational constraints yet still provides year-round coverage for phenological reasoning.

Spatially, the dataset's distribution (Fig.~\ref{fig:vqa-spatial-dist}) is concentrated in eastern and southwestern Australia. This alignment with the nation's primary agricultural zones and population centers ensures that the VQA tasks are grounded in areas of significant human activity and environmental relevance.

The distribution across the eight VQA categories (Fig.~\ref{fig:vqa-qa-chart}) is remarkably well-balanced, with each category comprising between 10.9\% and 14.0\% of the total question set. This deliberate balancing ensures a fair and robust evaluation of a VLM's capabilities across a diverse range of reasoning skills, preventing overall performance metrics from being skewed by any single task. 

The dataset's composition by satellite sensor (Figure~\ref{fig:vqa-sensor-chart}) highlights its multi-generational nature, a key feature for long-term Earth observation analysis. While the more recent Landsat 8 satellite serves as the primary data source (38.4\%), the dataset includes substantial contributions from Landsat 7 (24.5\%), Landsat 9 (22.9\%), and the historical Landsat 5 mission (14.2\%). This deliberate inclusion of four distinct sensors ensures that models are exposed to the full range of instrumental variations inherent in the Landsat program. Such diversity is crucial for developing robust models capable of performing consistent, long-term analysis across different eras of satellite technology.

The thematic content, visualized in the word cloud (Fig.~\ref{fig:vqa-wordcloud}), is dominated by terms related to land cover (\texttt{vegetation}, \texttt{water}, \texttt{cropland}), visual analysis (\texttt{visible}, \texttt{spatial}, \texttt{relationship}), and question-answering tasks (\texttt{type}, \texttt{dominant}, \texttt{matches}). This vocabulary underscores the dataset's focus on interpreting environmental features and complex spatial arrangements within Landsat imagery.

\subsection{Comparison with Remote-Sensing VLM Datasets}

\paragraph{Scope and diversity.}

For our comparative analysis, we selected prominent, large-scale, and open-source remote sensing VLM datasets that provide image-text pairs. While other notable datasets exist, some were excluded from a direct comparison of Landsat content due to a lack of relevant metadata. For instance, \textsc{RSICD} lacks image source information, Git-10M does not contain 30-meter GSD imagery, and \textsc{ChatEarthNet} is composed exclusively of Sentinel-2 data. Consequently, for the purposes of our Landsat-specific comparison, these datasets are considered to have zero relevant images. Our analysis therefore focuses on \textsc{ChatEarthNet}, \textsc{GAIA}, and \textsc{EarthDial}, which explicitly include Landsat data.

The comparison points in our analysis were derived as follows. The number of distinct Landsat satellites included and the presence of geo-location metadata were determined directly from the official descriptions and accompanying files for each dataset. However, the temporal span of the Landsat imagery was not always explicitly provided and required a methodical inference process. We established this span by using the earliest launch date of the included Landsat satellites as the start point and the dataset's public release date as the end point.

For example, \textsc{ChatEarthNet} and \textsc{GAIA} both include imagery from Landsat 8 (launched 2013). Given that \textsc{ChatEarthNet} was released in 2023 and \textsc{GAIA} in 2024, we infer their respective temporal spans to be 2013–2023 and 2013–2024. Similarly, since \textsc{EarthDial} exclusively contains Landsat 8 data and was released in 2024, its inferred span is also 2013–2024.

In stark contrast, the temporal span for \textsc{Landsat30-AU} is not an estimation. Because we retain precise, per-image metadata—including the specific satellite source and exact capture date for every image—we can definitively state our dataset's coverage from 1988 to 2024, a key advantage for reliable, long-term studies.

\paragraph{Linguistic and semantic richness.}

Our linguistic and semantic comparison focuses on the datasets that provide captions for their Landsat imagery: \textsc{ChatEarthNet} and \textsc{GAIA}. We evaluate two key properties: descriptive depth, measured by average caption length, and lexical diversity, using the Mean Segmental Type-Token Ratio (MSTTR). While \textsc{EarthDial} contains Landsat imagery, it is a VQA-focused dataset and lacks comparable long-form captions; consequently, it was excluded from our caption length and MSTTR analysis. We did include its VQA answers when calculating vocabulary size to offer a baseline for comparison.

The significant differences in the measured metrics are a direct result of each dataset's unique construction methodology. The captions in \textsc{ChatEarthNet} are generated automatically from OpenStreetMap (OSM) tags, a process that produces concise, object-centric descriptions and explains its very short average caption length of 9.3 words. In contrast, each image in the \textsc{GAIA} dataset is associated with five distinct captions. To facilitate a fair comparison, we concatenated these five descriptions into a single text entry for each image, which accounts for its substantially longer average caption length of 183.3 words. This approach operates under the reasonable assumption that the five captions for each image provide unique, non-duplicated information.

\section{Benchmark Evaluation}

\subsection{Task Settings.}

Our benchmark evaluation is structured around two primary tasks: \textbf{Image Captioning} and \textbf{VQA}. For the captioning task, models were provided only with the raw image. For the VQA task, models received the image, a question, and a set of multiple-choice options. To ensure a rigorous and objective evaluation, we enforced strict output handling rules: for captioning, the model's raw text output was evaluated without any cleaning or reformatting. For VQA, a response was considered correct only if it was an exact match to one of the provided options. All models are using the standardized prompts detailed in Listing~\ref{bechmark-ic-prompt} and Listing~\ref{bechmark-vqa-prompt}.

\subsubsection{Implementation Details.}

Our evaluation includes two categories of models: four specialized VLMs and four general VLMs. The specialized group consists of two remote sensing models (\textbf{EarthDial}, \textbf{RS-LLaVA}) and two reasoning models (\textbf{MiMo}, \textbf{GLM-V}), while the general group includes \textbf{Qwen}, \textbf{Llama}, \textbf{Gemma 3}, and \textbf{LLaVA}.

All models were evaluated under controlled conditions, receiving only the raw image as input, without any supplementary land cover or land use information. However, we established two distinct protocols to fairly assess the different model architectures.

The standard, non-reasoning VLMs were evaluated under identical, one-shot conditions. This approach ensures a direct and fair comparison of their performance on the benchmark tasks.

In contrast, the reasoning models were evaluated in a zero-shot setting to leverage their intrinsic chain-of-thought capabilities, with a maximum output limit of 8,192 tokens. Their generated output typically includes both intermediate reasoning steps and the final answer. To enable a fair comparison, we programmatically post-processed this output by stripping away the reasoning tokens. For the captioning task, only the resulting clean caption that fell within the token limit was evaluated. Similarly, for VQA tasks, we used a rigorous evaluation method. The final answer was extracted from the reasoning chain and compared against the ground truth with an exact match criteria.

In addition to evaluating the base models, we also fine-tuned Qwen and Llama using a parameter-efficient QLoRA scheme on a 15\% split of the combined \textsc{Landsat30-AU-Cap} and \textsc{Landsat30-AU-VQA} datasets. The model's backbone was first quantized to 4-bit NormalFloat. We then trained rank-64 LoRA adapters (with $\alpha=128$ and a dropout of 0.05) on the frozen, quantized weights. These adapters were inserted into all attention projections (q, k, v, o), the gated-MLP stack (gate, up, down), and the cross-modal vision projector. The models were trained for a single epoch. Optimization was performed using AdamW with a learning rate of $2 \times 10^{-4}$, cosine annealing, a 6\% warm-up fraction, and a weight decay of $10^{-6}$.

All experiments described were conducted on a server equipped with eight NVIDIA L4-24G GPUs.

\begin{listing*}[ht]%
\caption{Benchmark - Image-Captioning Prompt.}%
\label{bechmark-ic-prompt}%
\begin{lstlisting}[basicstyle=\ttfamily\small,breaklines=true,language=Python]
system_prompt = """You are an expert model for describing satellite or aerial images of landscapes, where each image pixel represents a 30-meter ground resolution. Use detailed, domain-specific language to describe the visible land covers, features, surface types (e.g., vegetation, artificial surfaces, water, etc.), and spatial relationships appropriate for the given spatial scale. Your goal is to give an analytical, objective caption that covers both the dominant and minor elements in the image, referencing spatial orientation (top-left, center, etc.) and notable connections (such as roads, patch boundaries, etc.). 
Base your descriptions only on observable features in the image, keeping the pixel resolution in mind."""

user_prompt = """Each image pixel corresponds to 30 meters on the ground. Respond in plain text only, with no formatting, lists, or special markup, just a single paragraph. Now, describe the following image in the same detailed manner, considering that each pixel represents 30 meters."""

\end{lstlisting}
\end{listing*}

\begin{listing*}[ht]%
\caption{Benchmark - VQA Prompt.}%
\label{bechmark-vqa-prompt}%
\begin{lstlisting}[basicstyle=\ttfamily\small,breaklines=true,language=Python]
system_prompt = """You are an evaluation agent for remote sensing VQA. Your ONLY job is to look at a satellite image, read the multiple choice question and its options, and pick exactly ONE best answer. The image pixel resolution is 30x30m.

Task
1. Inspect the image carefully.
2. Read the question and the list of answer options
3. Choose the single option that best answers the question, based solely on visual evidence.

Output rules
* Return only the text in current option.
* Do NOT output words, punctuation, or explanations.
* Trim whitespace;.

Example
(User supplies an image that clearly shows a branching network of channels entering a muddy coastline.)

Question:
Which land cover type is dominant in this image?

Options: ['Dense forest', 'Bare surface', 'Urban area', 'River delta'].

Answer: River delta"""

user_prompt = f"Question:{question_txt}\n\nOptions: {option_txt}\n"

\end{lstlisting}
\end{listing*}

\subsection{RQ1: How do Specialized VLMs perform compared to General models?}

Our analysis reveals that specialized models exhibit distinct and often conflicting performance profiles. To illustrate this, we first examine their factuality and tendency toward hallucination in the image captioning task. As shown in Table~\ref{tab:specialized_hallucination}, the reasoning VLM \textbf{GLM-V} stands out as the most factually grounded model, achieving the highest scores on both sentence-level (\textbf{1-CHAIR-s}) and instance-level (\textbf{1-CHAIR-i}) hallucination metrics. This is particularly noteworthy given its average caption length of 155 words; despite this verbosity, it maintains superior hallucination control compared to the more concise remote sensing models.

In contrast, the other reasoning model, \textbf{MiMo}, produces the longest captions, averaging 168 words, nearly 20\% longer than those of the remote sensing VLMs. This tendency toward verbosity correlates with a significantly higher hallucination rate, as it records the worst sentence-level score (0.3831) in this group. Meanwhile, the remote sensing VLMs, \textbf{EarthDial} and \textbf{RS-LLaVA}, demonstrate a more cautious approach. By generating shorter captions, they achieve strong sentence-level hallucination control, suggesting a shared, possibly inherent, design that prioritizes factual precision over descriptive detail.

\begin{table}[ht]
\centering
\small
\setlength{\tabcolsep}{4pt}
\begin{tabular}{@{} lcccc@ {}}
\toprule
\textbf{Model} & \textbf{1-CHAIR-s} & \textbf{1-CHAIR-i} & \textbf{Avg. Cap. Len.} \\
\midrule
\textsc{EarthDial}  & 0.5920 & 0.8197 & 140 \\
RS-LLaVA  & 0.5920 & 0.8119 & 139 \\
MiMo  & 0.3831 & 0.7805 & 168 \\
GLM-V  & \textbf{0.6259} & \textbf{0.8496} & 155 \\
\bottomrule
\end{tabular}
\caption{Hallucination and verbosity among Specialized VLMs. Bold indicates the best performance in each column.}
\label{tab:specialized_hallucination}
\end{table}

However, the strengths in captioning factuality do not translate to robust VQA performance. Table~\ref{tab:specialized_vqa_flaws} details the critical flaws observed in the specialized models. The remote sensing models, purportedly designed for this domain, show significant weaknesses. \textbf{EarthDial} fails on multiple reasoning categories (\textbf{APR}, \textbf{COA}, and \textbf{USR}), while \textbf{RS-LLaVA} performs surprisingly poorly on fundamental object recognition tasks (\textbf{SRI} and \textbf{USR}), achieving the lowest scores among all evaluated models. We hypothesize this is due to a domain mismatch between their remote sensing training data and our Landsat imagery benchmark. Conversely, the reasoning-VLM \textbf{MiMo} leverages its chain-of-thought capabilities to excel in numerical (\textbf{NUM}) and measurement (\textbf{MOP}) tasks, achieving the highest overall VQA score within this specialized group. \textbf{GLM-V}'s performance, while factually grounded in captions, remains unremarkable in VQA. This stark divergence underscores that no single specialized model provides consistent, all-around competence.

\begin{table}[ht]
\centering
\small
\setlength{\tabcolsep}{4pt}
\begin{tabular}{@{} lcccccc@ {}}
\toprule
\textbf{Model} & \textbf{APR} & \textbf{COA} & \textbf{NUM} & \textbf{MOP} & \textbf{SRI} & \textbf{USR} \\
\midrule
\textsc{EarthDial} & 0.2349 & 0.1034 & 0.4362 & 0.6116 & 0.5124 & 0.1552 \\
RS-LLaVA & \textbf{0.6857} & \textbf{0.8088} & 0.4985 & 0.6309 & 0.2617 & 0.1034 \\
MiMo & 0.4000 & 0.4577 & \textbf{0.6142} & \textbf{0.8430} & \textbf{0.9421} & \textbf{0.8897} \\
GLM-V & 0.4571 & 0.3636 & 0.5863 & 0.6749 & 0.6997 & 0.8828 \\
\bottomrule
\end{tabular}
\caption{VQA performance breakdown for Specialized VLMs. Bold indicates the best performance in each column.}
\label{tab:specialized_vqa_flaws}
\end{table}

Finally, to place these results in a broader context, Table~\ref{tab:overall_comparison} compares the top-performing specialized models against the best general models (without fine-tune). While \textbf{MiMo} shows competitive VQA abilities and \textbf{GLM-V} leads in factual captioning, they are ultimately outperformed in aggregate by generalist models. This finding is particularly noteworthy given the hypothesis that reasoning-focused VLMs would excel at remote sensing tasks, which often necessitate complex, multi-step logical deduction.

\begin{table}[ht]
\centering
\small
\setlength{\tabcolsep}{5pt}
\begin{tabular}{@{} llccc@ {}}
\toprule
\textbf{Type} & \textbf{Model} & \textbf{SPIDEr} & \textbf{1-CHAIR-s} & \textbf{VQA Overall} \\
\midrule
\multirow[t]{2}{*}{Specialized}
& MiMo & 0.0958 & 0.3831 & \textbf{0.7555} \\
& GLM-V & 0.1177 & \textbf{0.6259} & 0.6287 \\
\midrule
\multirow[t]{2}{*}{General}
& Qwen & 0.1114 & 0.4697 & 0.7428 \\
& Gemma 3 & \textbf{0.1246} & 0.3572 & 0.7356 \\
\bottomrule
\end{tabular}
\caption{Overall performance comparison of top specialized versus general  (without fine-tune). Bold indicates the best performance in each column.}
\label{tab:overall_comparison}
\end{table}

\subsection{RQ2: Can fine-tuning improve VLM performance in Landsat imagery understanding?}

Our analysis unequivocally demonstrates that fine-tuning provides a decisive performance boost for adapting general VLMs to Landsat imagery. We compare the base \textbf{Qwen} and \textbf{Llama} models against their fine-tuned counterparts, \textbf{Qwen-ft} and \textbf{Llama-ft}.

In the image captioning task, fine-tuning led to significant gains in semantic quality, as detailed in Table~\ref{tab:finetune_caption}. \textbf{Qwen-ft} established new state-of-the-art results, achieving the top score in \textbf{SPIDEr} (0.3054) while also improving its factual grounding, reflected by a higher \textbf{1-CHAIR-i} score. \textbf{Llama-ft} also saw a substantial 63\% improvement in its \textbf{SPIDEr} score; however, this came with a nuanced trade-off, as its resistance to object-level hallucination slightly decreased.

\begin{table}[ht]
\centering
\small
\setlength{\tabcolsep}{5pt}
\begin{tabular}{@{} lccc@ {}}
\toprule
\textbf{Model} & \textbf{SPIDEr} & \textbf{1-CHAIR-i} & \textbf{1-CHAIR-s} \\
\midrule
Qwen & 0.1114 & 0.7959 & 0.4697 \\
Qwen-ft & \textbf{0.3054} & \textbf{0.8549} & 0.4657 \\
\midrule
Llama & 0.1695 & 0.8296 & \textbf{0.5483} \\
Llama-ft & 0.2767 & 0.8016 & 0.5224 \\
\bottomrule
\end{tabular}
\caption{Impact of fine-tuning on image captioning performance. Bold indicates the best performance in each column.}
\label{tab:finetune_caption}
\end{table}

The most compelling evidence for the value of fine-tuning is found in the VQA results, where the models learned to address domain-specific challenges. As shown in Table~\ref{tab:finetune_vqa}, \textbf{Qwen-ft} more than doubled its accuracy on the \textbf{APR} task and achieved a perfect score on \textbf{FOD}, correctly learning the spatial resolution limits of the imagery. Overall, \textbf{Qwen-ft} achieved the highest VQA accuracy (0.8710) and secured the top score in six of the eight reasoning categories. These results confirm that even limited, efficient fine-tuning is a critical step for specializing VLMs to the unique visual and logical demands of Landsat imagery analysis.

\begin{table}[ht]
\centering
\small
\setlength{\tabcolsep}{5pt}
\begin{tabular}{@{} lccc@ {}}
\toprule
\textbf{Model} & \textbf{APR} & \textbf{FOD} & \textbf{Overall Acc.} \\
\midrule
Qwen & 0.2984 & 0.7167 & 0.7428 \\
Qwen-ft & \textbf{0.7016} & \textbf{1.0} & \textbf{0.8710} \\
\midrule
Llama & 0.3111 & 0.6633 & 0.6025 \\
Llama-ft & 0.5238 & \textbf{1.0} & 0.7315 \\
\bottomrule
\end{tabular}
\caption{VQA performance improvement after fine-tuning. Bold indicates the best performance in each column.}
\label{tab:finetune_vqa}
\end{table}

\subsection{RQ3: What are the strengths and weaknesses of VLMs on Landsat imagery?}

To identify the overarching strengths and weaknesses of current VLMs on Landsat imagery, we analyzed the per-category VQA accuracies across all evaluated models.

The results show that VLMs consistently excel at direct perceptual tasks that involve recognizing clear, unambiguous visual features. As detailed in Table~\ref{tab:vlm_strengths}, models like the fine-tuned \textbf{Qwen-ft} achieve near-perfect in identifying dominant land cover (\textbf{DLC}), confirming the presence of macro-objects (\textbf{MOP}), and correctly assessing the absence of sub-pixel features (\textbf{FOD}). This indicates that VLMs possess a strong baseline for grounded visual recognition in the remote sensing domain.

\begin{table}[ht]
\centering
\small

\setlength{\tabcolsep}{5pt}
\begin{tabular}{@{} lccc@ {}}
\toprule
\textbf{Model} & \textbf{DLC} & \textbf{MOP} & \textbf{FOD} \\
\midrule
Qwen-ft & \textbf{0.9651} & \textbf{0.8678} & \textbf{1.0} \\
MiMo & 0.9247 & 0.8430 & 0.9333 \\
Gemma 3 & 0.9220 & 0.7934 & 0.4533 \\
Qwen & 0.9409 & 0.7603 & 0.7167 \\
\bottomrule
\end{tabular}
\caption{VLM high performance on VQA tasks. Bold indicates the best performance in each column.}
\label{tab:vlm_strengths}
\end{table}

However, performance degrades significantly as tasks demand more abstract, contextual, or fine-grained reasoning. Table~\ref{tab:vlm_weaknesses} highlights these challenges. Numerosity (\textbf{NUM}) emerges as a universal bottleneck, with even the top-performing model, \textbf{Qwen-ft}, scoring only 0.6588. Tasks requiring holistic scene interpretation—such as judging urban scale (\textbf{USR}) or assessing cloud usability (\textbf{COA})—yield highly polarized results. For example, \textbf{Gemma 3} excels at \textbf{USR} (0.9310) while \textbf{RS-LLaVA} fails (0.1034). The most abstract tasks, like inferring seasonality from subtle textures (\textbf{APR}), remain difficult across the board and are primary beneficiaries of targeted fine-tuning. This pattern indicates that while current VLMs excel at direct perception, the next frontier is developing their capacity for complex, multi-layered reasoning on satellite imagery. A dedicated satellite imagery source, such as Landsat, could be instrumental in advancing this capability.

\begin{table}[ht]
\centering
\small
\setlength{\tabcolsep}{5pt}
\begin{tabular}{@{} lcccc@ {}}
\toprule
\textbf{Model} & \textbf{NUM} & \textbf{USR} & \textbf{COA} & \textbf{APR} \\
\midrule
Qwen-ft & 0.6588 & 0.8966 & 0.9530 & 0.7016 \\
MiMo & 0.6142 & 0.8897 & 0.4577 & 0.4000 \\
Gemma 3 & \underline{0.3234} & 0.9310 & 0.8150 & 0.6730 \\
RS-LLaVA & 0.4985 & \underline{0.1034} & 0.8088 & 0.6857 \\
\bottomrule
\end{tabular}
\caption{VLM performance bottlenecks on abstract and contextual reasoning tasks. \underline{Underlined scores} denote particularly poor performance.}
\label{tab:vlm_weaknesses}
\end{table}

\subsection{RQ4: How does fine-tuning affect model robustness across different sensors?}

To assess model robustness to sensor-specific variations, we breakdown the VQA test set results by the source Landsat satellite (Landsat 5, 7, 8, and 9). We compare the performance consistency of base models against their fine-tuned counterparts across these four sensors.

As shown in Table~\ref{tab:sensor_vqa_metric}, base models exhibit significant performance variance across different sensors, indicating a lack of robustness to shifts in sensor characteristics. For instance, the standard Llama model's accuracy fluctuates from 0.670 on Landsat 8 down to 0.541 on Landsat 5. This inconsistency highlights a critical weakness for real-world applications, where a model must perform reliably on imagery from any available satellite.

Fine-tuning on \LandsatAU{} proves crucial for mitigating this issue. After fine-tuning, Llama-ft's performance stabilizes dramatically, with scores tightly clustered between 0.717 and 0.743 across all four sensors. This alignment reduces the standard deviation in performance across sensors from 0.057 to just 0.010, a testament to improved robustness. Similarly, while the base Qwen model struggles with Landsat 9 imagery (0.679 accuracy), fine-tuning boosts its performance on this sensor to 0.910, bringing it in line with its performance on other sensors. This demonstrates that training on \LandsatAU{}, with its diverse, multi-sensor composition, is critical for mitigating sensor-specific biases and developing models that can generalize reliably across the entire 36-year Landsat archive.

\begin{table*}[ht]
\centering
\small
\setlength{\tabcolsep}{5pt}
\begin{tabular}{@{}llccccccc@{}}
\toprule
Type & Model & Size & LS5 (ls5t)$\uparrow$ & LS7 (ls7e)$\uparrow$ & LS8 (ls8c)$\uparrow$ & LS9 (ls9c)$\uparrow$ & Overall & Std Dev ($\sigma \downarrow$) \\
\midrule
\multirow[t]{4}{*}{Specialized}
& EarthDial & 4B & 0.520 & 0.453 & 0.514 & 0.444 & 0.483 & 0.041 \\
& RS-LLaVA & 7B & 0.580 & 0.619 & 0.491 & 0.646 & 0.572 & 0.068 \\
& MiMo & 7B & 0.761 & 0.721 & \textbf{0.808} & 0.707 & 0.756 & 0.046 \\
& GLM-V & 9B & 0.684 & 0.622 & 0.657 & 0.559 & 0.629 & 0.058 \\
\midrule
\multirow[t]{4}{*}{General}
& Qwen & 7B & 0.732 & 0.722 & 0.802 & 0.679 & 0.743 & 0.054 \\
& LLaVA & 8B & 0.588 & 0.578 & 0.681 & 0.543 & 0.610 & 0.062 \\
& Llama & 11B & 0.541 & 0.589 & 0.670 & 0.548 & 0.602 & 0.059 \\
& Gemma 3 & 12B & 0.782 & 0.743 & 0.712 & 0.738 & 0.736 & 0.029 \\
\midrule
\multirow[t]{2}{*}{General with ft}
& Qwen-ft & 7B & \textbf{0.895} & \textbf{0.919} & 0.805 & \textbf{0.910} & \textbf{0.871} & 0.052 \\
& Llama-ft & 11B & 0.717 & 0.733 & 0.743 & 0.720 & 0.731 & \textbf{0.012} \\
\bottomrule
\end{tabular}
\caption{VQA performance breakdown by Landsat sensor. The scores represent the accuracy for each model on the subset of the VQA test set corresponding to each satellite. Fine-tuning significantly improves performance consistency across sensors, as indicated by the lower standard deviation ($\sigma$). Bold indicates the best performance in each column.}
\label{tab:sensor_vqa_metric}
\end{table*}